\pdfoutput = 1
\documentclass[twocolumn]{article}

\usepackage{times}
\usepackage[numbers]{natbib}
\usepackage[english]{babel}
\usepackage{blindtext}
\usepackage{graphicx}
\usepackage{amsmath, amsthm, amssymb, bbm, bm}
\usepackage{enumerate}
\usepackage{float}      
\usepackage{subcaption}  
\usepackage{wrapfig}
\usepackage[margin=0cm]{caption}
\usepackage[titletoc, toc]{appendix}
\usepackage{tabularx}

\usepackage{multirow}
\usepackage{hhline}
\usepackage{makecell}
\usepackage{placeins}  

\usepackage[x11names, usenames, dvipsnames, svgnames, table]{xcolor}
\definecolor{firebrick}{rgb}{.698,.133,.133}
\definecolor{mybluelight}{rgb}{0.9, 0.9, 1.}
\definecolor{myorangelight}{rgb}{1., 0.9, 0.9}

\usepackage[utf8]{inputenc} 
\usepackage[T1]{fontenc}    
\usepackage{url}            
\usepackage{booktabs, colortbl}       
\usepackage{amsfonts}       
\usepackage{nicefrac}       
\usepackage{microtype}      

\usepackage{csquotes}
\usepackage{latexsym}

\usepackage{pifont}
\usepackage[boxruled, vlined, linesnumbered]{algorithm2e}
\SetAlFnt{\small}
\SetAlCapFnt{\small}
\SetAlCapNameFnt{\small}
\usepackage{algorithmic}
\algsetup{linenosize=\tiny}

\let\oldnl\nl
\newcommand{\nonl}{\renewcommand{\nl}{\let\nl\oldnl}}

\usepackage{paralist}

\usepackage{xspace}
\usepackage{soul}
\usepackage{dsfont}
\usepackage{stmaryrd}
\usepackage[textwidth=15mm]{todonotes}
\usepackage{dirtytalk}
\usepackage{pbox}
\usepackage{cprotect}

\usepackage{verbatim}
\usepackage{textcomp}
\usepackage[normalem]{ulem}

\usepackage{mathtools}
\usepackage{etextools}
\usepackage[inline]{enumitem}

\usepackage[colorlinks=true,allcolors=firebrick,bookmarks=false,breaklinks=true]{hyperref}

\newcommand\pxap{\texttt{PxAP}\xspace}

\definecolor{darkergreen}{RGB}{21, 152, 56}
\definecolor{red2}{RGB}{252, 54, 65}
\definecolor{Gray}{gray}{0.85}
\newcolumntype{g}{>{\columncolor{Gray}}c}

\let\OLDthebibliography\thebibliography
\renewcommand\thebibliography[1]{
  \OLDthebibliography{#1}
  \setlength{\parskip}{0pt}
  \setlength{\itemsep}{0pt plus 0.3ex}
}

\DeclareMathOperator*{\argmax}{argmax}

\theoremstyle{definition}

\DeclarePairedDelimiterX{\divx}[2]{(}{)}{%
  #1\;\delimsize\|\;#2%
}

\usepackage{style}

\newcommand\cl{\texttt{CL}\xspace}

\makeatletter
\@namedef{ver@everyshi.sty}{}
\newcommand{\removelatexerror}{\let\@latex@error\@gobble}

\newcommand\glas{\texttt{GlaS}\xspace}
\newcommand\camsixteen{\texttt{CAMELYON16}\xspace}

\newcommand\beloc{\texttt{B-LOC}\xspace}
\newcommand\becl{\texttt{B-CL}\xspace}

\title{Source-Free Domain Adaptation of Weakly-Supervised Object Localization Models for Histology}

\renewcommand\footnotemark{}

\author{
  Alexis~Guichemerre$^{1}$,
  ~Soufiane~Belharbi$^{1}$,
  ~Tsiry~Mayet$^{2}$,
  ~Shakeeb~Murtaza$^{1}$,
  ~Pourya~Shamsolmoali$^{3}$,
  \\~\textbf{Luke~McCaffrey}$^{4}$,
  ~\textbf{Eric~Granger}$^{1}$\\
 $^1$ LIVIA, Dept. of Systems Engineering, ETS Montreal, Canada\\
 $^2$ LITIS UR 4108, F-746000, INSA Rouen Normandie, France\\
 $^3$ EEECS, Queen's University Belfast, UK\\
$^4$  Goodman Cancer Research Centre, Dept. of Oncology, McGill University, Montreal, Canada\\
{\tt\footnotesize \textcolor{black}{alexis.guichemerre.1@ens.etsmtl.ca} }
}

\newcommand{\ignore}[1]{}

\begin{document}
\maketitle\thispagestyle{fancy}

\maketitle
\rhead{\color{gray} \small Guichemerre et al. \;  [CVPRw 2024]}

\begin{abstract}
Given the emergence of deep learning, digital pathology has gained popularity for cancer diagnosis based on histology images. Deep weakly supervised object localization (WSOL) models can be trained to classify histology images according to cancer grade and identify regions of interest (ROIs) for interpretation, using inexpensive global image-class annotations.  A WSOL  model initially trained on some labeled source image data can be adapted using unlabeled target data in cases of significant domain shifts caused by variations in staining, scanners, and cancer type.  
In this paper, we focus on source-free (unsupervised) domain adaptation (SFDA), a challenging problem where a pre-trained source model is adapted to a new target domain without using any source domain data for privacy and efficiency reasons.  
SFDA of WSOL models raises several challenges in histology, most notably because they are not intended to adapt for both classification and localization tasks.  
In this paper, 4 state-of-the-art SFDA methods, each one representative of a main SFDA family, are compared for WSOL in terms of classification and localization accuracy. They are the SFDA-Distribution Estimation, Source HypOthesis Transfer, Cross-Domain Contrastive Learning, and Adaptively Domain Statistics Alignment. 
Experimental\footnote{Code: \href{https://github.com/AlexisGuichemerreCode/survey_hist_wsol_sfda}{https://github.com/AlexisGuichemerreCode/}\allowbreak\href{https://github.com/AlexisGuichemerreCode/survey_hist_wsol_sfda}{survey\_hist\_wsol\_sfda}.} results on the challenging Glas (smaller, breast cancer) and Camelyon16 (larger, colon cancer) histology datasets indicate that these SFDA methods typically perform poorly for localization after adaptation when optimized for classification. 
\end{abstract}

\textbf{Keywords:} Source Free Unsupervised Domain Adaptation, Weakly-supervised Object Localization, Histology Images, Class Activation Mapping (CAM), Deep Models.
%
%

\section{Introduction}
\label{sec:intro}

Histology images obtained by microscopy of biopsy tissue play a crucial role in cancer diagnosis. These images are valuable resources for pathologists, allowing them to assess tumour type and aggressiveness \citep{HE2012538}, and potentially adapt the patient's treatment.  Identifying cellular features or tumour markers remains a challenging and costly undertaking because it relies on a pathologist's expertise and experience \citep{komura2018machine,subtypes2022pathology,rony23}. Given the intra- and inter-variability of diagnoses, this strong dependence on the pathologist makes reproducibility of grading challenging \citep{komura2018machine}. 

Recent methods from the machine learning (ML) and computer vision communities can assist the pathologist in the diagnosis of cancers based on histology images \citep{fan2022}. Early image analysis methods for digital pathology rely on ML models such as thresholding \citep{Thresholding_techniques}, edge detection \citep{woods2008}, post-processing\citep{woods2008}, and clustering.  
The advent of deep learning (DL) has enabled the development of more robust models for classification, regression, segmentation, and localization to handle the complexity and variability of histology images. Despite the effectiveness of DL models like convolutional neural networks (CNNs) and vision transformers (ViTs), they are often seen as "black boxes" \citep{rony23, komura2018machine}. This is problematic for medical decision-making, where understanding the model's decision is crucial for supporting the pathologist's analysis. 

In histology, whole slide images (WSIs) are captured at a very high resolution (over 200 million pixels) \citep{rony23}. Extracting pixel-level annotations for supervised training of a segmentation model is costly and time-consuming. However, WSOL models can provide spatial visualization linked to a classifier's predictions after training on images sampled from WSIs annotated with inexpensive image-class labels. Given an input image, WSOL models can predict the cancer grade and identify ROIs. However, these models should be adapted using unlabeled target data when applied to histology images captured under different conditions (e.g., staining, scanners, and cancer type) to address domain shifts.

The performance of WSOL models typically declines due to the shift in data distribution between the training (source) and testing (target) domains \citep{farahani2021brief, kouw2019review, wang2018deep}. Several discrepancy-based and adversarial methods have been proposed for unsupervised domain adaptation (UDA) of ML and DL models, using labeled images from a source domain and unlabeled images from the target domain \citep{farahani2021brief, kouw2019review, wang2018deep}.   
Given the privacy, confidentiality, and logistical challenges, source data may not be available for medical imaging applications. To overcome these limitations, source-free (unsupervised) domain adaptation (SFDA) methods have been proposed, where source data is not required during the adaptation process \citep{fang2024source, yu2302comprehensive, zhang2023source}. SFDA is challenging since labeled source data cannot be used during adaptation to align source and target distributions. 

This paper focuses on white-box SFDA methods, divided into two main categories: data generation and fine-tuning \citep{fang2024source}, as illustrated in Fig. \ref{fig:tax}).  
Regarding data generation, some strategies employ style transfer to adjust target images to imitate the source domain's style \citep{hou2021source}. Other methods aim to enrich the target domain with labeled images by generating content in the target style\citep{kurmi2021domain}. \citep{qiu2021sourcefree, ding2022sourcefree, tian2021vdmda} aim to create features similar to those of the source domain to align features between the target and source domains.  \citep{wang2022cross} propose using contrastive learning, where positive samples are pulled close while negative ones are pushed apart. In the absence of the source domain, the classifier's weights may be used as prototypes of each class learned on the source domain.  \citep{yang2021generalized,yang2021exploiting} focus more on using the neighbours to guide the adaptation process for each image. Finally, teacher-student models also exploit criteria based on information maximization and entropy minimization \citep{yang2021transformerbased, chen2021selfsupervised}. Although SFDA methods have been applied extensively for classification of natural images \citep{fang2024source, yu2302comprehensive, zhang2023source}, their performance for WSOL (classification and localization) of histology images has not been established. 

In this paper, state-of-the-art SFDA methods are empirically evaluated for the adaptation of three WSOL models: Deep MIL \citep{deepmil}, GradCAM++ \cite{gradcampp} and TS-CAM \citep{tscam}. The four SFDA methods we compare are representative of the two main white-box SFDA families -- SFDA-Distribution Estimation (SFDA-DE) \citep{ding2022sourcefree} based on domain distribution generation, Source HypOthesis Transfer (SHOT) \citep{yang2021transformerbased} based on hidden structure mining, Cross-Domain Contrastive Learning (CDCL) \citep{wang2022cross} based on contrastive learning, and Adaptively Domain Statistics Alignment (AdaDSA) \citep{fan2022ieee} based on domain alignment via statistics. Experiments are conducted on two challenging public datasets -- GlaS (small size, colon cancer) and Camelyon512 (large size breast cancer) -- and SFDA methods are compared in terms of classification and localization accuracy.


\section{Related Work}

\subsection{Weakly Supervised Object Localization}
WSOL aims to localize object instances in an image by utilizing only class-level labels. Class Activation Mapping (CAM) is a widely used approach in WSOL, to extract the ROIs of a particular object based on the spatial feature maps of a CNN~\cite{zhou2016learning}. However, CAM-based methods focus on discriminant image regions and produce low-resolution maps. To overcome this issue, \citep{deepmil} proposes DeepMIL, which uses a multi-instance learning framework to identify objects in images by first constructing instance representations, then computing a global representation of the bag (image) using an attention mechanism that weights these instances, with strong attention weights indicating regions of interest (ROIs) and weak weights indicating background. Also gradient-based methods have been proposed to highlight resolution maps by fusing the upsampled CAMs with the gradient of output class w.r.t. input image~\cite{ChattopadhyaySH18wacvgradcampp, SelvarajuCDVPB17iccvgradcam}. Grad-CAM++ \cite{gradcampp} builds on these methods by computing weights from higher-order derivatives of the class output for the feature maps, allowing for more accurate and class-discriminating visualizations. This method refines the resolution maps by accounting for the pixel-wise impact on classification decisions.  

Since models used to obtain CAMs are only trained using image-class labels, they struggle to cover less discriminative regions. This issue has been addressed by data augmentation~\cite{choe2019attention, wei2017object, yun2019cutmix, zhang2018adversarial, singh2017hide}, where the network is encouraged to explore less discriminative regions of a particular object. For instance, the most popular method removes the most discriminative regions and enables the network to look beyond those regions through adversarial perturbation~\cite{singh2017hide, yun2019cutmix} or using adversarial loss~\cite{choe2019attention,zhang2018adversarial}. Other methods focus on directly producing localization maps instead of post-processing spatial information obtained from internal activations of network~\cite{rahimi2020pairwise, xue2019danet, zhang2020rethinking}. 

Despite substantial improvements, these methods may still highlight background regions. To address this issue, \cite{wu2021background, zhu2021background} propose to suppress background regions so the network can identify foreground regions corresponding to a particular class with a high level of confidence. 
\begin{figure*}[!t]
    \centering
    \includegraphics[width=\linewidth]{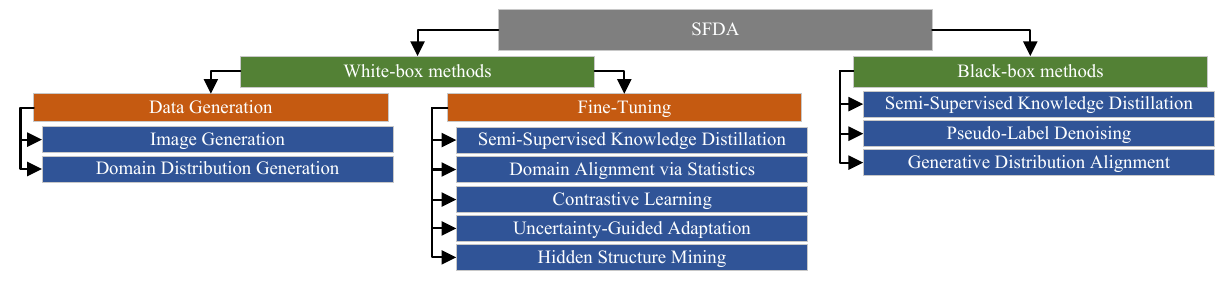}
    \caption{Overall taxonomy of SFDA methods for classification as defined in \citep{fang2024source}.}
    \label{fig:tax}
\end{figure*}
Separate modules have been proposed to optimize localization and classification accuracy by decomposing object parts for classification and localization tasks, ensuring that the model can retain comparable classification accuracy~\cite{meng2021foreground}. To regularize the internal CNN features for capturing different object parts,  \cite{xie2021online} added encoder-decoder modules into different network layers, thereby preserving object details at different levels. Additionally, \cite{zhang2020inter, pan2021unveiling} are proposed to expand foreground maps to expand maps to the edges of the concerned object. Despite significant advancements, these models struggle to capture long-range within different object parts, causing partial activation over different object parts. To overcome this issue, different transformer-based methods have been proposed to expand activation over different object parts \cite{bai2022weakly, chen2022lctr, gupta2022vitol, su2022re, li2022caft, meng2022adversarial, tscam}. TS-CAM~\cite{tscam} was the first method that employed a transformer for WSOL by displacing the class head from the class token to the patch token, and by combining activations presented in the classification head with class tokens. However, this introduces background noise into the maps. To address this issue, \cite{bai2022weakly} proposes a calibration module in the transformer for producing smooth activation across different object parts. \\ Despite their impressive results, these methods are very sensitive to threshold values. Recently, different methods have been shown to generate fine-grained localization maps while learning from noisy pseudo-labels harvested from activations of different networks~\cite{murtaza2022dips, murtaza2022dipssypo, Murtaza2023dips, belharbi2022fcam}. A domain adaptation-based technique has also been proposed to improve localization accuracy~\cite{zhu2022weakly}.

\subsection{Source-Free Domain Adaptation}

SFDA methods can be categorized into white-box and black-box families \citep{fang2024source} (see Fig. \ref{fig:tax}). White-box methods access the information contained in the source network, while black-box methods only use the output of the source network for adaptation. White-box models can be further classified into two categories: data generation and fine-tuning methods \citep{fang2024source}.

The data generation family of white-box SFDA methods is based on generating images in either the source style  \citep{hou2021source, yang2022source} or the target style \citep{kurmi2021domain, Li2020cvpr}. Concerning image generation in source style, \citep{hou2021source} uses a generator to modify target images in the source style using the features of the original and generated images. Style is transferred by matching the mean and variance values of the features in each backbone layer to the mean and variance values contained in the batch normalization layers of the source network. As a result, the source network should be able to process this image to classify it correctly. This idea is also adopted in \citep{ yang2022source}, using the same technique to generate images, but using these images to obtain features that are close to those of the source domain. These features are then used for domain alignment between source and target. 

\citep{kurmi2021domain, Li2020cvpr} consider generating images in the target style. \citep{kurmi2021domain} train a conditional generator to produce images based on the desired class. The source network is used to guarantee the class, and then a first discriminator receives the generated image and an image from the target domain to assure consistency of style and content of the generated image wrt the target domain. The feature extractor of the target network is trained such that another discriminator can generate similar features between the generated image and the target domain images. The feature extractor is also trained to produce the same output for the generated image.

The fine-tuning family of white-box SFDA methods seeks to adapt the feature extractor such that it produces features for the target domain that are similar to those of the source. \citep{ding2022proxymix} considers using target images close to the anchor in the feature space to create a surrogate source domain. Another possibility proposed by several authors is to generate source-like features \citep{qiu2021sourcefree, tian2021vdmda, ding2022sourcefree}. \citep{qiu2021sourcefree} propose to generate features via a generator trained with the weight of the classifier and contrastive learning. In \citep{tian2021vdmda}, the authors propose to model a virtual domain based on a Gaussian Mixture Model (GMM) composed of a Gaussian density per class. The hypothesis of estimating the distribution of the source by a Gaussian is also adopted by \citep{ding2022sourcefree}. The images of the target domain with the pseudo-labels closest to the anchors (from the spherical $k$-means) are stored and used to estimate the means and variances of each Gaussian. With the Gaussian modeling and the ability to generate source-like features, alignment is achieved through Contrastive Domain Discrepancy (CDD) as introduced in \citep{kang2019contrastive}.

In line with this idea of generating the same features for each class, \citep{Adversarial_contrastive,wang2022crossdomain,agarwal2021unsupervised} propose methods based on contrastive learning. The intention is to guide the network to learn similar representations for the same class, close together in the feature space while pushing away different samples. In \citep{wang2022crossdomain}, the anchors used for the contrastive loss are the weights of the source classifier. As a result, the loss function aligns features with the same pseudo-label obtained by a clustering method with the anchors for the same class.

Instead of using data generation, some SFDA methods are based on the concept of knowledge distillation \citep{yu2022sourcefree, liu2021graph, liu2022ieeets, yang2021transformerbased, chen2021selfsupervised}. These methods adapt using a teacher-student architecture. Initially, both networks are initialized with the source model and the teacher model is used to guide the student model through the learning process. The teacher network is smoothly updated with the exponential moving average (EMA). Therefore, at each adaptation stage, the teacher network incorporates more information from the target domain. In \citep{yang2021transformerbased}, the teacher model is used to guide the student network using the pseudo-labeling obtained from the teacher features. In \citep{yu2022sourcefree}, the authors divides the target domain
images into two subgroups based on the cross entropy loss function and use the teacher network to guide the "noisy" images. Also, other work extends the approach with the use of multiple teachers and students to build more robust models using information from each source network. 

An important part of fine-tuning methods is the pseudo-labeling images. To mitigate misclassifications, some approaches rely on a Monte Carlo process that passes an image through the network multiple times with various dropouts. The purpose is to reduce the difference between the prediction with and without dropouts. These methods are usually used in addition to other loss functions to enhance the model. For example, in \citep{hegde2021uncertaintyaware}, a teacher-student approach is used, and the Monte Carlo process is added to the predictions of the teacher model to improve it. 

Other SFDA methods focus on the relationships between multiple images in the feature space, rather than on the image itself. The idea is to assume that similar feature representations must potentially belong to the same class. In \citep{yang2021generalized}, the network is trained to encourage the predictions of each image over its neighbours using stored predictions. In \citep{yang2021exploiting}, two sets are defined, containing feature representations and prediction scores to compute the $k$ nearest neighbours for each target image. They add an affinity parameter for each determined neighbor to strengthen the relationship.

Another strategy for SFDA consists of using the information contained in the batch normalization (BN) layers of the source network. In \citep{ishii2021sourcefree}, they use information maximization combined with a loss function that adjusts the distribution between the target feature output by the encoder and the distribution contained in the classifier's BN layer using Kullback–Leibler divergence. \citep{liu2021miccai} proposes a more progressive adaptation of mean and variance information in batch normalization layers. The scaling and shifting parameters are used such that they must remain close to those of the source, as they are higher-order features. In \citep{fan2022ieee}, the impact of the source is smoothly added during the adaptation process through a weight.

\begin{figure*}[!h]
    \centering
    \includegraphics[width=\linewidth]{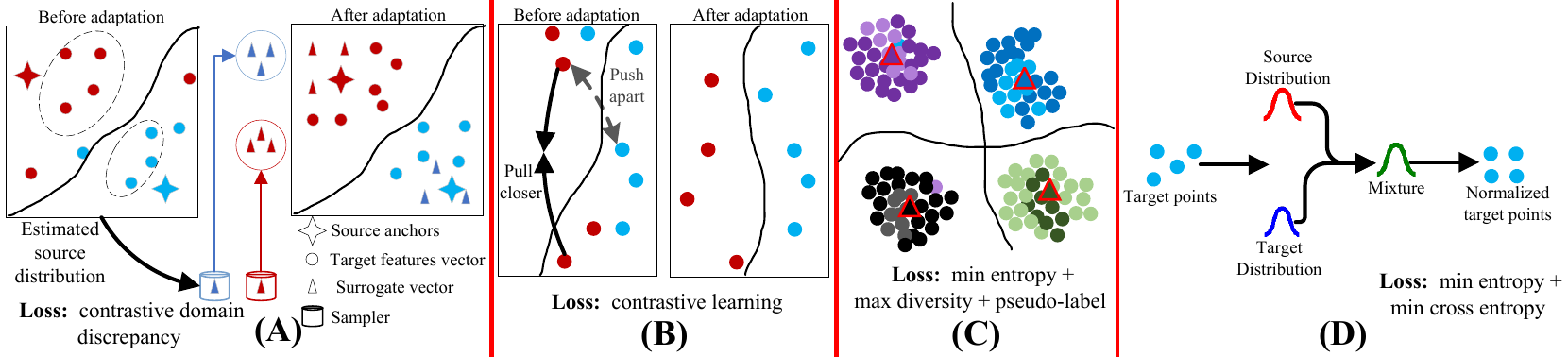}
    \caption{Illustration of SFDA process. (A) The SFDA-DE method is based on distribution estimation. It generates features in the style of the source to perform alignment with target data. (B) CDCL is based on contrastive learning, where positive samples are pulled close while negative ones are pushed apart. In the absence of the source domain, the classifier's weights are used to define prototypes of each class learned on source data. (C) SHOT is a hidden structure mining method based on information maximization. (D) AdaDSA focuses on combining the statistics of the BN layers from the source to normalize target data.}
    \label{fig:main-comp}
\end{figure*}

\section{Comparative Analysis of SFDA Methods}

Initially, a neural network noted $\phi_{source}=g.h$ is trained on a source domain $S=\{x_{s},y_{s}\}_{i=1}^{n_s}$, where $y_{s}$ belongs to the set of $K$ classes and $g: X \rightarrow \mathbb{R}^d$ is the feature extractor and  $h: \mathbb{R}^d \rightarrow \mathbb{R}^K$ is the classifier where $w^G \in \mathbb{R}^{m \times K}$ denote the weights learned and $w^{G}_{k} \in \mathbb{R}^m$ is the $k$-th weight vector of $w^G$. In SFDA, the source domain is only used for pre-training the source network. Then, the goal is to train a target network $\phi_{target}$ using only the unlabeled target data $T=\{x_{t}\}$, where $x_{t} \in \mathcal{X}_t$ and $t= {1,2,...,n}$. 

This section provides additional details on the 4 white-box SFDA methods compared in this paper to adapt WSOL models for histology. Each SFDA method represents a family of SFDA methods -- either data generation or fine-tuning.  They are the SFDA-DE \citep{ding2022sourcefree} based on domain distribution generation, SHOT \citep{yang2021transformerbased} based on hidden structure mining, CDCL \citep{wang2022cross} based on contrastive learning, and AdaDSA \citep{fan2022ieee} based on domain alignment via statistics. 

\noindent \textbf{(A) SFDA-DE:} 
\citep{ding2022sourcefree} propose an approach to estimate the source class distribution with a Gaussian distribution combined with target features information. Firstly, pseudo-labels $\hat{y}^t$ are estimated by applying spherical $k$-means. Only a subset of samples is preserved, based on distance criterion to the corresponding anchor $A_{\hat{y}^t_i}$ to avoid adding samples with incorrect pseudo-labels for the Gaussian estimation. 
Given the restricted number of reliable features, it is possible to estimate the parameters of each Gaussian more accurately for each class $k$ according to:
\begin{equation}
\mathcal{N}_k^{sur}({\|\bar{f_k^t}\|}_2\frac{\mathbf{w}_k^G}{{\|\mathbf{w}_k^G\|}_2},  \frac{\gamma\cdot\textbf{f}_k^{\,t}\cdot{\textbf{f}_k^{\,t}}^\top}{\sum\limits_{x_i^t \in \mathcal{D}_t'}\mathds{1}(\hat{y}_i^t=k)}), k\in\mathcal{C},
\label{sur_dis}
\end{equation}
where $\mathbbm{1}_{y =k}$ is defined as the indicator function and $\textbf{f}$ the target features. Consequently, for each class, it is possible to estimate the distribution and thus generate source-like features. The adaptation is achieved through Contrastive Domain Discrepancy (CDD) based on Maximum Mean Discrepancy (MMD) and a subset of the class  $\mathcal{C}'$:
\begin{equation}
\mathcal{L}_{\text{CDD}}=\frac{~\sum\limits_{k\in\mathcal{C}'}\mathcal{L}_{\text{MMD}}^{k,k} }{|\mathcal{C}'|} - \frac{~\sum\limits_{k_1\in\mathcal{C}'}\sum\limits_{k_2\in\mathcal{C}'}^{k_1\neq k_2}\mathcal{L}_{\text{MMD}}^{k_1,k_2}}{|\mathcal{C}'|(|\mathcal{C}'|-1)},
\label{cddloss}
\end{equation}

By estimating the source distribution, we gain a better understanding of the distribution that our target features should adopt, facilitating the efficient use of the classifier. However, the method proposed the disadvantage that it relies essentially on the pseudo-labels process. The accuracy of these pseudo-labels will drastically impact the effectiveness of the approach. A higher precision in the estimation of those labels allows us to estimate the Gaussian distribution more accurately and also push the corresponding feature to the right direction in the feature space at the opposite high level on the incorrect pseudo label will add to much error in the estimation of the Gaussian and the target features will not be pushed in a correct direction.

\noindent \textbf{(B) CDCL:} Contrastive learning methods are designed to encourage the network to learn representations that make similar samples (belonging to the same class) closer in the feature space, while dissimilar samples are pushed away. As with the SFDA-DE method \citep{ding2022sourcefree}, a clustering method based on spherical $k$-means is achieved to obtain pseudo-labels at the beginning of each epoch. The pseudo-labels are essential as they allow creation of positive pairs and negative pairs for the contrastive loss defined as follows:
\begin{equation}
  \mathcal{L}_{CDCL} = - \sum\limits_{k=1}^{K} \mathbf{1}_{\hat{y}^i_t = k} \log \frac{\exp ({f_{t}^{i}}^\top {\bm w}_s^k / \tau) }{\sum\limits_{j=1}^{K} \exp ({f_{t}^{i}}^\top {\bm w}_s^j / \tau)}.
  \label{eqn:cdc_sdf}
\end{equation}
where $\tau$ is the temperature parameter. Unlike SFDA-DE, CDCL presents the advantage of not estimating source feature distribution, and only relying on using classifier weights which reduces the impact of adding error in the adaption process. Also, the contrastive loss helps to separate the features for each class in the feature space. However, like SFDA-DE, this approach relies on the accuracy of pseudo-labels derived from spherical $k$-means.

\noindent \textbf{(C) SHOT:} 
The previous methods use pseudo-labels from spherical $k$-means to guide the network. These techniques can be limited in domain adaptation as they rely only on the precision of the labels. In SHOT, they propose to use an information maximization loss. This loss is composed of an entropy and a diversity loss. This combination allows to improve the confidence in its prediction but also encourages the last one to consider all the classes. 
\begin{equation}
    \footnotesize \mathcal{L}_{ent}(\phi_{t};\mathcal{X}_t) =    -\mathbb{E}_{x_t\in\mathcal{X}_t} \sum\nolimits_{k=1}^{K} \delta_k(\phi_{t}(x_t)) \log \delta_k(\phi_{t}(x_t)),
\end{equation}
\begin{equation}
\footnotesize
\mathcal{L}_{div}(\phi_{t};\mathcal{X}_t) = \sum\nolimits_{k=1}^{K} \hat{p}_k \log \hat{p}_k 
    = D_{KL}(\hat{p}, \frac{1}{K}\mathbf{1}_K) - \log K,
\label{eq:ent}
\end{equation}
Where $\hat{p}_k$ is the mean output embedding of the whole target domain
The information maximization loss plays a crucial role in SHOT since it allows a clear separation between classes in the feature space. This aspect is important because SHOT also relies on a cross-entropy loss based on pseudo-labels $\hat{y}_t$ obtained with a self-supervised pseudo-labeling strategy inspired by $k$-means:
\begin{equation}
    \begin{aligned}
        &\mathcal{L}_{self} = \ \mathbb{E}_{(x_t,\hat{y}_t)\in \mathcal{X}_t \times \hat{\mathcal{Y}}_t } \sum\nolimits_{k=1}^{K} \mathds{1}_{[k=\hat{y}_t]} \log \delta_k(\phi_{t}(x_t)),
    \end{aligned}
    \label{eq:overall}
\end{equation}    
Eq. \ref{eq:overall} encourages the network to focus on more than its prediction with the $\mathcal{L}_{ent}$. SHOT has the advantage of not depending primarily on pseudo-labels. Information maximization is composed of entropy and divergence loss. Entropy loss can be beneficial if the source model produces correct outputs in the early stage and the divergence loss allows for the separation of the classes which is helpful for the self-supervising pseudo-label strategy. Nevertheless, the entropy loss is ineffective if it forces the network to be more confident in its predictions, even if they are incorrect, while the pseudo labels may be more accurate.

\noindent \textbf{(D) AdaDSA:} 
The latest method we've explored is based on Batch Normalization information. The statistics from each Batch Normalization (BN) layer are used to estimate the distribution of source data and use this information to perform cross-domain adaptation by smoothly updating the values of the mean and variance of each BN layer. This moment of the cross-domain adaptation is done as follows:
\begin{equation}
\begin{aligned}
\mu_{ts} &= \alpha \mu_{t} + (1 - \alpha) \mu_{s} \\
\sigma_{ts} &= \alpha (\sigma_{t}^{2} + (\mu_{t} - \mu_{t}^{s})^{2}) + (1 - \alpha)(\sigma_{s}^{2} + (\mu_{s} - \mu_{t}^{s})^{2})
\end{aligned}
\end{equation}
where $\mu_{t}$, $\mu_{s}$,$\sigma_{t}$ and $\sigma_{s}$ are the first and second moment of each BN layer on Target and Source. $\alpha$ is a parameter that controls the influence of the weights of each domain in the cross-domain adaptation. Note that the model is trained using a classic entropy loss to ensure confidence in model predictions. A cross-entropy loss was also used to avoid forgetting information contained in the source network by using the label $y'_t$ defined as:
\begin{equation}
    \begin{aligned}
        y'_t = \argmax_{y} \left\{ (1 - \lambda) \phi_s(x_t)[y] + \lambda \phi_t(\alpha; x_t)[y] \right\}
    \end{aligned}
\end{equation}

\section{Results and Discussion}

\subsection{Experimental Methodology:}

\paragraph{GlaS dataset.}  
This dataset is used for colon cancer diagnosis. The dataset is composed of 165 images from 16 Hematoxylin and Eosin (H\&E) and contains labels at both pixel-level and image-level (benign or malign). The dataset consists of 67 images for training, 18 for validation and 80 for testing.  We use the same protocol as in \citep{rony23}, i.e. 3 examples per class fully supervised for \beloc selection.

\paragraph{CAMELYON dataset.} 
A patch-based benchmark is derived from the Camelyon16 dataset that contains 399 Whole slide images with two classes (normal and metastasic) for the detection of metastases in H\&E-stained tissue sections of sentinel lymph nodes (SLNs) from women with breast cancer. Patch extraction of size ${512\times 512}$ follows a protocol established by \citep{rony23} to obtain patches with annotations at image and pixel level. 
The benchmark contains a total of 48870 images, including 24348 for training, 8850 for validation and 15664 for testing. From the validation dataset, 6 examples per class are randomly selected to be fully supervised to determine \beloc as defined in \citep{rony23}.

\paragraph{Implementation details.}
For pretraining on the source domain, we use the same setup as defined in \citep{rony23}. We use the pretrained source network to initialize the target network. For SFDA-DE, CDCL, and SHOT, the classifier is frozen during the SFDA stage. The hyperparameters for the different WSOL models and  all SFDA methods are defined in the appendix.

\paragraph{Experimental measures.} 
To compare different models, we use two measures: standard classification accuracy to assess classifier performance and localization accuracy, measured by PxAP, to generate localization masks from activation maps. We apply various thresholds, normalizing the activation maps via min-max normalization to identify the thresholds that optimize performance, with a precise mathematical definition of the average accuracy per pixel as defined in \citep{choe2020evaluating, rony23}. The true positive, false negative, false positive, and true negative rates are defined in \citep{fpr}.

\setlength{\tabcolsep}{3pt}
\renewcommand{\arraystretch}{0.6}
\begin{table*}[ht]
\caption{Performance of SFDA methods on \camsixteen $\rightarrow$ \glas and \glas $\rightarrow$ \camsixteen test set. Performance measures are the true positive, false negative, false positive, and true negative rates.  Model selection is performed on \beloc for the source model.}
\vspace{-4pt}
\centering
\resizebox{\textwidth}{!}{%
\centering
\small
\begin{tabular}{lcc*{4}{c}c*{4}{c}c*{4}{c}c*{4}{c}}
& &  & \multicolumn{8}{c}{\camsixteen $\rightarrow$ \glas}  & & \multicolumn{8}{c}{\glas $\rightarrow$ \camsixteen} \\
\cline{4-12}\cline{14-22} \\
 & \textbf{Methods} &  & tpr & fnr & fpr & tnr  &  & tpr & fnr & fpr & tnr &  & tpr & fnr & fpr & tnr &  & tpr & fnr & fpr & tnr\\
 \hline \\
\multirow{15}{*}{\rotatebox[origin=c]{90}{Deep MIL \citep{deepmil}}} & Source only  &  & 60.4 & 39.5 & 62.9  & 37.0  &  & 60.4 & 39.5 & 62.9  & 37.0 &  & 23.9 & 76.0 & 84.5  & 15.4 &  & 23.9 & 76.0 & 84
5& 15.4 \\
\cline{2-2}\cline{4-7}\cline{9-12}\cline{14-17}\cline{19-22} \\
& \textbf{SFDA}  &  &  & &  &  &  & &  &  &  & &  &  &  && &  & & &   \\
\cline{2-2}\cline{4-7}\cline{9-12}\cline{14-17}\cline{19-22} \\
& SFDA-DE \citep{ding2022sourcefree} {\small \emph{(cvpr,2022)}}  &  & 60.4  & 39.5  & 62.9  & 37.0  &  & 51.6 & 48.3 & 67.0  & 32.9 &  & 48.6 & 51.3 & 83.7  & 16.2 &  & 23.9 & 76.0 & 84.5 & 15.4 \\
& SHOT \citep{yang2021transformerbased}{\small \emph{(icml,2020)}} &  & 60.4  & 39.5  & 62.9  & 37.0 &  & 41.8 & 58.1 & 61.0  & 38.9 &  & 68.5 & 31.4 & 75.8  & 24.1 &  & 0.0 & 99.9 & 99.9  & 0.0 \\
& CDCL \citep{wang2022cross} {\small \emph{(tmm,2022)}} &  & 68.5 & 31.4  & 54.2  & 45.7  &  & 68.5 & 31.4 & 54.2  & 45.7 &  & 50.9 & 49.0 & 86.6  & 13.3 &  & 10.21 & 89.7 & 88.5  & 11.4 \\
& AdaDSA \citep{fan2022ieee} {\small \emph{(tsnre,2022)}} &  & 29.6  & 70.3  & 74.8  & 25.1  &  & 58.9& 41.0 & 35.1  & 64.8 &  & 25.7 & 74.2 & 80.1  & 19.8 &  & 24.8 & 75.1 & 83.5  & 16.4 \\

\cline{2-2}\cline{4-7}\cline{9-12}\cline{14-17}\cline{19-22} \\
& \textbf{Fine-tuning}  &  & \multicolumn{17}{c}{}  \\
\cline{2-2}\cline{4-7}\cline{9-12}\cline{14-17}\cline{19-22} \\
& Image-Net \citep{ImageNet} {\small \emph{(neurips,2012)}} &  & 63.3  & 36.6  & 76.3  & 23.6  &  & 58.1 & 41.8 & 65.4  & 34.5 &  & 66.9 & 33.0 & 89.5  & 10.42 &  & 66.1 & 33.3 & 90.3  & 9.6 \\ 
\cline{2-2}\cline{4-7}\cline{9-12}\cline{14-17}\cline{19-22} \\
\hline \\
& Source only  &  & 61.0  & 38.9  & 35.9  & 64.0  &  & 61.0  & 38.9  & 35.9  & 64.0 &  & 45.1 & 54.8 & 82.9  & 17.0 &  & 45.1 & 54.8 & 82.9  & 17.0 \\
\cline{2-2}\cline{4-7}\cline{9-12}\cline{14-17}\cline{19-22} \\
\multirow{12}{*}{\rotatebox[origin=c]{90}{GradCAM++ \citep{gradcampp}}} & \textbf{SFDA}  &  &  & &  &  &  & &  &  &  & &  &  &  && &  & & &   \\
\cline{2-2}\cline{4-7}\cline{9-12}\cline{14-17}\cline{19-22} \\
& SFDA-DE \citep{ding2022sourcefree} {\small \emph{(cvpr,2022)}} &  &  61.08  & 38.9  & 35.9  & 64.0  &  &  40.4  & 59.53  & 56.3  & 43.6 &  &65.0 & 34.9 & 84.9  & 15.0 &  & 37.9 & 62.0 & 80.8  & 19.1 \\
& SHOT \citep{yang2021transformerbased} {\small \emph{(icml,2020)}} &  & 82.7  & 17.2  & 36.6  &  63.3  &  & 52.8 & 47.1 & 51.7  &48.2  &  & 82.7 & 17.2 & 36.6  & 63.3 &  & 0.7 & 99.2 & 99.1  & 0.8 \\
& CDCL \citep{wang2022cross} {\small \emph{(tmm,2022)}} &  & 67.7 & 32.2  & 41.7  & 58.2  &  & 56.3 & 43.6 & 56.9  & 43.0 &  & 68.9 & 31.0 & 80.9  & 19.0 &  & 45.1 & 54.8 & 82.9  & 17.0 \\
& AdaDSA \citep{fan2022ieee} {\small \emph{(tsnre,2022)}} &  & 29.7  & 70.2  & 82.9  & 17.0  &  & 28.5 & 71.4 & 77.3  & 22.6 &  & 48.2 & 51.7 & 84.3  & 15.6 &  & 47.3 & 52.6 & 84.4  & 15.5 \\

\cline{2-2}\cline{4-7}\cline{9-12}\cline{14-17}\cline{19-22} \\
& \textbf{Fine-tuning}  &  & \multicolumn{17}{c}{}  \\
\cline{2-2}\cline{4-7}\cline{9-12}\cline{14-17}\cline{19-22} \\
& Image-Net \citep{ImageNet} {\small \emph{(neurips,2012)}}&  & 62.0  & 37.9  & 79.8  & 20.1  &  & 59.3 & 40.6 & 84  & 15.9 &  & 42.1 & 57.8 & 89.4  & 10.5 &  & 11.3 & 88.6 & 91.5  & 8.43  \\
\cline{2-2}\cline{4-7}\cline{9-12}\cline{14-17}\cline{19-22} \\
\hline \\
& Source only  &  & 42.2  & 57.7  & 52.7 & 47.2  &  & 42.2  & 57.7  & 52.7 & 47.2 &  & 0.0 & 100.0 & 100.0  & 0.0 &  & 0.0 & 100.0 & 100.0  & 0 \\
\cline{2-2}\cline{4-7}\cline{9-12}\cline{14-17}\cline{19-22} \\
\multirow{12}{*}{\rotatebox[origin=c]{90}{TS-CAM \citep{tscam}}} & \textbf{SFDA}  &  &  & &  &  &  & &  &  &  & &  &  &  && &  & & &   \\
\cline{2-2}\cline{4-7}\cline{9-12}\cline{14-17}\cline{19-22} \\
& SFDA-DE \citep{ding2022sourcefree} {\small \emph{(cvpr,2022)}} &   & 69.9  & 30.0  & 40.7  & 59.2 &  & 67.6 & 32.8 & 43.2  & 56.7 &  & 74.1 & 25.8 & 75.2  & 24.7 &  & 0.0 & 100.0 & 100.0  & 0.0 \\
& SHOT \citep{yang2021transformerbased}{\small \emph{(icml,2020)}} &   & 69.8  & 30.1  & 42.4  & 57.5  &  & 68.5& 31.4 & 41.1  & 58.8 &  & 30.7 & 69.2 & 89.6  & 10.3 &  & 50.0 & 49.9 & 73.0  & 26.9 \\
& CDCL  \citep{wang2022cross} {\small \emph{(tmm,2022)}} &   & 52.3  & 47.6  & 55.7  & 44.2  &  & 71.6 & 28.3 & 42.3  & 57.6 &  & 44.5 & 55.4 & 81.0 & 18.9 &  & 0.0 & 100.0 & 100.0 & 0.0 \\
& AdaDSA \citep{fan2022ieee}{\small \emph{(tsnre,2022)}} &  & 42.2 & 57.7  & 52.7  & 47.2 &  & 42.2 & 57.7 & 52.7  & 47.2 &  & 0.0 & 100.0 & 100.0  & 0.0 &  & 0.0 & 100.0 & 100.0  & 0.0 \\

\cline{2-2}\cline{4-7}\cline{9-12}\cline{14-17}\cline{19-22} \\
& \textbf{Fine-tuning}  &  & \multicolumn{17}{c}{}  \\
\cline{2-2}\cline{4-7}\cline{9-12}\cline{14-17}\cline{19-22} \\
& Image-Net \citep{ImageNet} {\small \emph{(neurips,2012)}}&   & 67.7  & 32.3  & 62.2  & 37.7  &  & 69.4 & 30.5 & 58.5  & 41.6&  & 70.2 & 29.7 & 73.2  & 26.7 &  & 53.3 & 46.6 & 76.5  & 23.4  \\
\cline{2-22} \\ \\
&\textbf{Fully supervised}  &  & \multicolumn{17}{c}{}  \\
\hline \\
& U-net \citep{unet}{\small \emph{(miccai,2015)}} &  & 88.9  & 11.0  & 89.7  & 10.2  &  & n/a &n/a &n/a  & n/a &  & 68.0 & 31.9 & 94.5  & 5.4&  & n/a & n/a & n/a  & n/a \\
\hline \\ 
\end{tabular}
}
\label{tab:mtx-conf-best-loc}
\vspace{-1em}
\end{table*}

\setlength{\tabcolsep}{3pt}
\renewcommand{\arraystretch}{0.8}
\begin{table*}[htbp]
\caption{Localization (\pxap) and classification (\cl) accuracy with model selection methods: \beloc/\becl on \glas and \camsixteen test sets. \beloc selection for the source network.}
\vspace{-4pt}
\centering 
\resizebox{0.55\linewidth}{!}{
\small
\begin{tabular}{l  c  c c c c c c c}
& &  & \multicolumn{2}{c}{\camsixteen $\rightarrow$ \glas}  & & \multicolumn{2}{c}{\glas $\rightarrow$ \camsixteen} \\
\cline{4-5}\cline{7-8} \\
 & \textbf{Methods} &  & \pxap & \cl &  & \pxap & \cl \\
 \hline \\
  &  Source only &  & 65.5 & 67.5 &  & 29.0 & 55.2 \\  
  \cline{2-2}\cline{4-5}\cline{7-8} \\
\multirow{8}{*}{\rotatebox[origin=c]{90}{Deep MIL \citep{deepmil} }} & \textbf{SFDA}  &  & & &  &  & \\
\cline{2-2}\cline{4-5}\cline{7-8} \\

& SFDA-DE \citep{ding2022sourcefree}{\small \emph{(cvpr,2022)}}&  &66.3 / 64.7&81.2 / 87.5& &46.1 / 29.0 & 51.8 / 55.2& \\
& SHOT \citep{yang2021transformerbased}{\small \emph{(icml,2020)}} &   &65.5 / 52.1&67.5 / 76.2& &45.6 / 25.3&51.3 / 67.5&  \\
& CDCL \citep{wang2022cross}{\small \emph{(tmm,2022)}} &  &66.3 / 66.3&81.2 / 81.2& &53.2 / 22.8&53.0 / 60.9& \\
& AdaDSA \citep{fan2022ieee} {\small \emph{(tsnre,2022)}} &  &56.1 / 47.6&46.2 / 46.2& &27.3 / 29.0&54.3 / 57.9& \\
\cline{2-2}\cline{4-5}\cline{7-8} \\
&\textbf{Fine-tuning}  &  & & &  &  & \\
\cline{2-2}\cline{4-5}\cline{7-8} \\

& Image-Net \citep{ImageNet}&  &79.9 / 50.0 &100.0 / 100.0 & &71.3 / 60.6 &85.0 / 89.9& \\
\hline \\

\multirow{12}{*}{\rotatebox[origin=c]{90}{GradCAM++ \citep{gradcampp}}} &  Source only &  &52.9 & 53.7 &  & 39.7 & 52.4 \\
\cline{2-2}\cline{4-5}\cline{7-8} \\
 & \textbf{SFDA}  &  & & &  &  & \\
 \cline{2-2}\cline{4-5}\cline{7-8} \\
&SFDA-DE \citep{ding2022sourcefree}{\small \emph{(cvpr,2022)}} &  &54.6 / 51.1&53.7 / 63.7& &55.4 / 33.0 & 49.7 / 56.4& \\
& SHOT \citep{yang2021transformerbased}{\small \emph{(icml,2020)}} &  &59.8 / 54.6&58.7 / 85.0& &47.1 / 18.8&50.5 / 65.5& \\
& CDCL  \citep{wang2022cross}{\small \emph{(tmm,2022)}}&  &54.6 / 56.3&53.7 / 62.5& &53.1 / 39.7&51.5 / 52.4& \\
& AdaDSA \citep{fan2022ieee}{\small \emph{(tsnre,2022)}}&  &59.9 / 53.7&46.2 / 46.2& &44.1 / 43.8&53.1 / 53.2& \\
\cline{2-2}\cline{4-5}\cline{7-8} \\
&\textbf{Fine-tuning}  &   &  & &  &  & \\
\cline{2-2}\cline{4-5}\cline{7-8} \\

& Image-Net \citep{ImageNet} {\small \emph{(neurips,2012)}}&  &76.8 / 77.9&100.0 / 100.0& &49.1 / 21.6 &63.4 / 89.4 & \\
\hline \\

\multirow{12}{*}{\rotatebox[origin=c]{90}{TS-CAM \citep{tscam}}} &  Source only &  &48.3 & 47.5 &  & 15.0 & 50.7& \\
\cline{2-2}\cline{4-5}\cline{7-8} \\
 & \textbf{SFDA}  &  & & &  &  & \\
 \cline{2-2}\cline{4-5}\cline{7-8} \\
&SFDA-DE \citep{ding2022sourcefree}{\small \emph{(cvpr,2022)}} &  &53.4/ 52.8 &66.2 / 68.7 &  & 47.8 / 15.8& 50.0 / 52.7&  \\
& SHOT \citep{yang2021transformerbased}{\small \emph{(icml,2020)}} &  &54.8 / 52.7 & 56.2 / 68.7 &  & 32.7 / 32.7 & 48.3 / 58.6& \\
& CDCL  \citep{wang2022cross} {\small \emph{(tmm,2022)}}&  &55.4 / 54.0 & 51.2 / 55.0 &  & 36.7 / 20.4 & 52.4 / 52.8&  \\
& AdaDSA \citep{fan2022ieee}{\small \emph{(tsnre,2022)}}&  & 48.3 / 48.3 & 47.5 / 47.5 &  & 15.0 / 15.0 & 50.7 / 50.7& \\
\cline{2-2}\cline{4-5}\cline{7-8} \\
&\textbf{Fine-tuning}  &  & & &  &  & \\
\cline{2-2}\cline{4-5}\cline{7-8} \\

& Image-Net \citep{ImageNet} {\small \emph{(neurips,2012)}}&  &65.4 / 63.5& 97.5 / 97.5& &41.6 / 35.1&51.4 / 88.4& \\
\hline \\
& U-net \citep{unet}{\small \emph{(miccai,2015)}} &  &95.8& n/a& &81.6&n/a& \\
\hline \\
\end{tabular}
}
\label{tab:accuracy-bloc-source}
\vspace{-1em}
\end{table*}

\setlength{\tabcolsep}{2pt}
\renewcommand{\arraystretch}{1.19}
\begin{table}[htbp]
\caption{Localization (\pxap) and classification (\cl) accuracy with model selection methods: \beloc / \becl on \glas test sets. \becl selection for the source network.}
\vspace{-4pt}
\centering
\resizebox{0.7\linewidth}{!}{%
\small
\begin{tabular}{l c  c@{\hspace{4mm}} c c}
& && \multicolumn{2}{c}{\camsixteen $\rightarrow$ \glas} \\
\cline{4-5} \\
& \textbf{Methods} && \pxap & \cl \\
\hline \\
& Source only && 64.5  & 81.2 \\
\cline{2-2} \cline{4-5} \\
\multirow{8}{*}{\rotatebox[origin=c]{90}{Deep MIL \citep{deepmil}}} & \textbf{SFDA}  &  &\\
\cline{2-2} \cline{4-5} \\

& SFDA-DE \citep{ding2022sourcefree} {\small \emph{(cvpr,2022)}}&&  65.9 / 64.5  & 53.7 / 81.2 \\
& SHOT \citep{yang2021transformerbased}{\small \emph{(icml,2020)}} && 64.5 / 55.7 & 81.2 / 86.2 \\
& CDCL \citep{wang2022cross}{\small \emph{(tmm,2022)}} && 65.9 / 64.5  & 70.0 / 81.2 \\
& AdaDSA \citep{fan2022ieee}{\small \emph{(tsnre,2022)}} && 49.4 / 48.9  & 46.2 / 46.2 \\
\cline{2-2} \cline{4-5} \\
& \textbf{Fine-tuning}  &  & \\
\cline{2-2} \cline{4-5} \\
&Image-Net \citep{ImageNet} {\small \emph{(neurips,2012)}}&& 79.9 / 50.0 &100.0 / 100.0\\
\hline \\

\multirow{12}{*}{\rotatebox[origin=c]{90}{GradCAM++ \citep{gradcampp}}} &  Source only && 56.1  & 75.0 \\
\cline{2-2} \cline{4-5} \\
& \textbf{SFDA}  &  & \\
\cline{2-2} \cline{4-5}  \\
&SFDA-DE\citep{ding2022sourcefree} {\small \emph{(cvpr,2022)}} && 70.0 / 55.1 & 82.5 / 97.5\\
& SHOT \citep{yang2021transformerbased}{\small \emph{(icml,2020)}} && 68.5 / 51.3  & 60.0 / 97.5 \\
& CDCL \citep{wang2022cross}{\small \emph{(tmm,2022)}}&& 67.8 / 54.7 & 85.0 / 95.0 \\
& AdaDSA \citep{fan2022ieee}{\small \emph{(tsnre,2022)}}&& 50.6 / 49.1  & 46.2 / 46.2 \\
\cline{2-2} \cline{4-5} \\
& \textbf{Fine-tuning}  &  & \\
\cline{2-2} \cline{4-5} \\
&Image-Net \citep{ImageNet} {\small \emph{(neurips,2012)}}&& 76.8 / 77.9&100.0 / 100.0\\
\hline \\

\multirow{12}{*}{\rotatebox[origin=c]{90}{TS-CAM \citep{tscam}}} &  Source only && 48.7 & 61.2 \\
\cline{2-2} \cline{4-5} \\
& \textbf{SFDA}  &  & \\
\cline{2-2} \cline{4-5} \\
&SFDA-DE \citep{ding2022sourcefree}{\small \emph{(cvpr,2022)}} && 53.9 / 53.9 & 46.2 / 71.2  \\
& SHOT \citep{yang2021transformerbased}{\small \emph{(icml,2020)}} && 53.8 / 51.1  & 66.2 / 90.0\\
& CDCL \citep{wang2022cross} {\small \emph{(tmm,2022)}}&&  48.7 / 55.9  & 53.7 / 70.0 \\
& AdaDSA \citep{fan2022ieee} {\small \emph{(tsnre,2022)}}&&  48.7 / 48.7 & 61.2 / 61.2 \\
\cline{2-2} \cline{4-5} \\
& \textbf{Fine-tuning}  &  & \\
\cline{2-2} \cline{4-5} \\
&Image-Net \citep{ImageNet} {\small \emph{(neurips,2012)}}&&65.4 / 63.5& 97.5 / 97.5\\
\hline \\
& U-net \citep{unet}{\small \emph{(miccai,2015)}} && 95.8 & n/a \\
\hline \\
\end{tabular}
}
\label{tab:accuracy-bcl-source}
\vspace{-1em}
\end{table}

\subsection{Impact of SFDA on Localization:} 

In Table ~\ref{tab:accuracy-bloc-source}, we observe improvements in classification accuracy for the methods DeepMil, GradCAM++, and TS-CAM when applied to the SFDA approaches SFDA-DE, SHOT, and CDCL. However, AdaDSA encounters challenges in model adaptation. The results presented in Table ~\ref{tab:accuracy-bloc-source} show a clear mismatch between the accuracy of localization and classification in the context of SFDA. When selecting models based on \becl for DeepMIL on the GlaS dataset, the results underscore the challenges of enhancing both localization and classification simultaneously. However, when \beloc is used as the criterion after adapting models on GlaS, improvements are observed in both localization and classification for methods like GradCAM++ and TS-CAM. Despite these improvements, it is important to note that using \beloc as the selection criterion after adaptation results in slightly lower classification accuracy compared to \becl. This reduction in classification accuracy can be attributed to the conflicting demands of localization and classification tasks, as discussed in \citep{rony23}. Although choosing \beloc over \becl leads to a minor decrease in classification accuracy, it significantly enhances the performance of localization tasks, highlighting a trade-off between optimizing for one task over the other.

In contrast to the GLAS dataset, applying the CAMELYON dataset showcases the challenges in balancing localization and classification tasks, particularly when \beloc is selected. This selection underlines SFDA methods' difficulties in optimizing classification. Most models, including DeepMil, GradCam++, and TS-CAM, did not show improvement in classification accuracy with \beloc, except in two instances: AdaDSA for GradCam++ and CDCL for TS-CAM, as shown in Table ~\ref{tab:accuracy-bloc-source}.

This is also observed in Tab.~\ref{tab:accuracy-bcl-source} when selecting \becl for the source model on CAMELYON. Based on our findings, choosing \beloc instead of \becl after the adaptation led to an important degradation in classification accuracy (-37\% by using SHOT with GradCAM++, -25\% using SFDA-DE with TS-CAM and -27.5\% using SFDA-DE with DeepMIL). Tab.~\ref{tab:accuracy-bcl-source} shows that while classification accuracy decreases, localization accuracy increases, indicating a trade-off between these tasks. This difference should be an important consideration for SFDA for histology. As illustrated in Tab.~\ref{tab:mtx-conf-best-loc} when GlaS is the target, the true positive and true negative rates are significantly improved by selecting \beloc for GradCAM++ with SHOT as an example (+29.9\% tpr, -34.4\% fnr, -15.1\% fpr, +15.1\% tnr) and for DeepMIL with SFDA-DE  (+8.8\% tpr, -8.8\% fnr, -4.1\% fpr, 4.1\% tnr).

\subsection{\beloc or \becl Selection for Source Networks:} 

In WSOL, early stopping is performed to select the best localization or classification models on validation data. The two models obtained after training are antagonistic, meaning that the best localization model is obtained in the first training epochs, while the best classification model is obtained in the last \citep{rony23}. A consequence of this property is that the model obtained for the best localization is less reliable for the classification. We can observe this analysis in Tab.~\ref{tab:accuracy-bloc-source} (fine-tuning: ImageNet). For the CAMELYON dataset, it is important to highlight the significant improvement in classification accuracy for GradCAM++ and TS-CAM, with increases of +36\% and +37\%, respectively.

In SFDA, the process of model selection is important. Specifically, methods like SFDA-DE, SHOT, and CDCL depend on generating pseudo-labels through $k$-means or Spherical $k$-means clustering. This generated information is then utilized to steer the network's learning process. Therefore, selecting \becl or \beloc for the source is an important criterion to consider for the adaptation. In Tabs.~\ref{tab:accuracy-bloc-source} and~\ref{tab:accuracy-bcl-source}, it is possible to compare the results obtained using \beloc or \becl on CAMELYON for the source and analyze the impact on the final models obtained using GLAS. It is observed that before adaptation, the source model chosen based on \becl consistently achieves better performance on the target domain (GLAS). This is expected, as \becl is designed with a stronger focus on classification tasks compared to \beloc, thereby outperforming it as a source model selection requirement.

As mentioned, several methods rely on clustering techniques at the beginning of each epoch to obtain pseudo-labels and guide the network during the adaptation process. If the source model is more efficient at classifying target images, the clustering task will be easier and the adaptation will be less challenging. We observe the effect of \beloc and \becl in the adaptation process as shown in Table ~\ref{tab:accuracy-bloc-source} and ~\ref{tab:accuracy-bcl-source} for GradCAM++ and TS-CAM with a gain for each SFDA methods ( GradCAM++ : + 33.8\%, +12.5\%, +32.5\% +0.0\% ; TS-CAM :  +2.5\%, +21.3\%, +15\%, + 13.7\%).

We did not perform the adaptation from GlaS to CAMELYON using the \becl method on GlaS, as the initial performance of GlaS as the source dataset already reaches a 100\% accuracy. Technically, using this model is the best we can get considering GlaS as source (100 \% accuracy). However, the results obtained without any adaptation are limited. The classification accuracy is 55.2\%, 52.4\% and 50.7\% for DeepMil, GradCAM++ and TS-CAM, respectively. Therefore, especially for challenging data like CAMELYON, the starting point of the fitting model is not ideal. Since the classification accuracy is low on the target domain (CAMELYON) highlights the difficulty for a source model trained on a not challenging to correctly classify new images. 
A larger source domain size tends to greatly improve the adaptation process when \becl is favoured over \beloc, resulting in a positive overall impact on performance.
\begin{figure}[!htb]
    \centering
    \includegraphics[width=\linewidth]{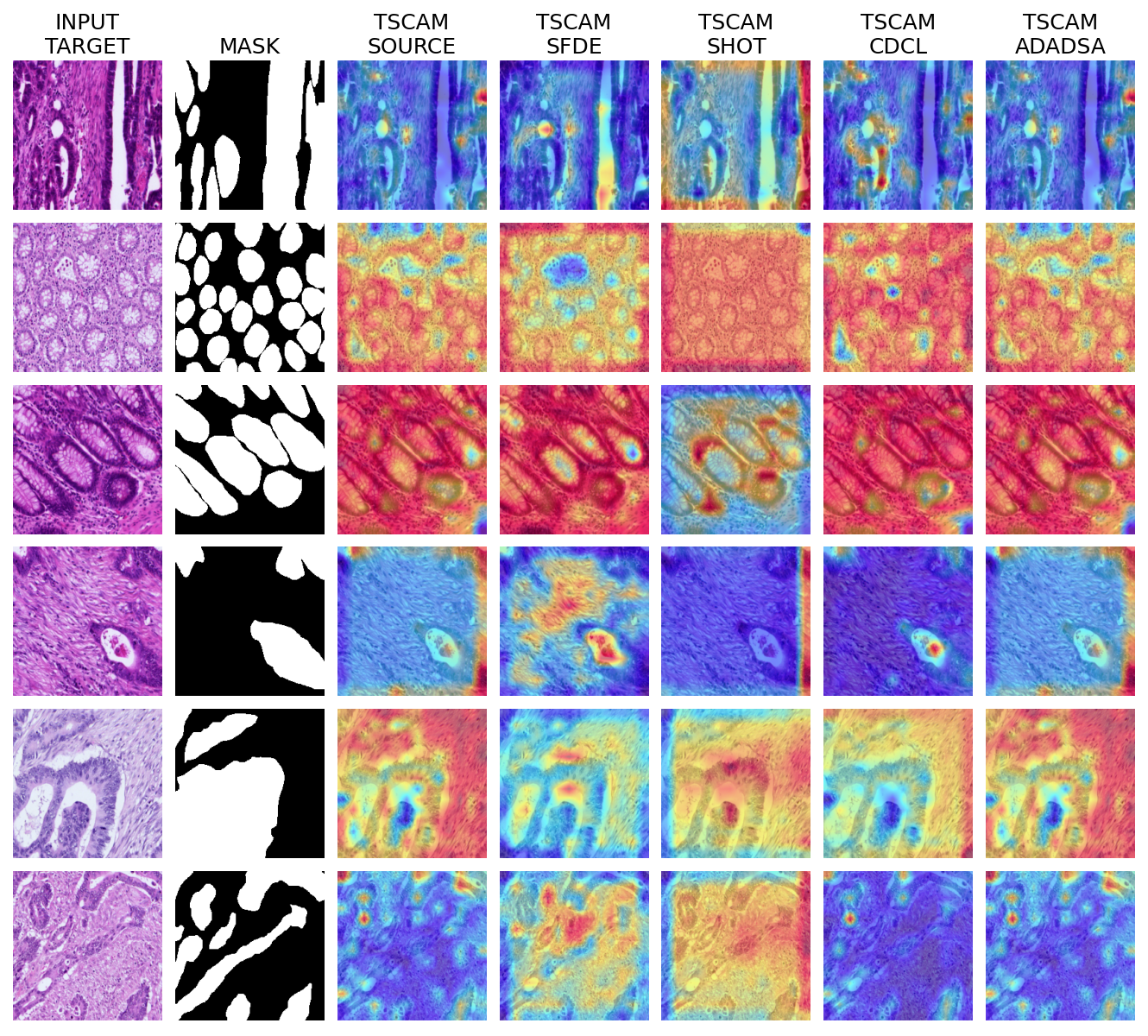}
    \caption{Visualisation on target (GLAS) dataset with TS-CAM with best localization source model trained with CAMELYON with source's best classification.}
\end{figure}

\subsection{Impact of Source Network Training:} 

The selection of the source model is a key component for the success of SFDA. We analyzed that the best setup to consider when adapting to GlaS is to use \becl as the source model for CAMELYON. The performance for GradCAM++ and TS-CAM are high that when using \beloc on GlaS as illustrated in Table~\ref{tab:accuracy-bloc-source} and~\ref{tab:accuracy-bcl-source}. Nevertheless, this is not established for DeepMIL as we can state. The classification accuracy is not necessarily improved by using \becl for the source and \becl after the adaptation with SFDA-DE, CDCL and AdaDSA (-6.3\%, +0.0\%, +0.0\%) but increases for SHOT (+10\%). The results obtained for DeepMil can be explained by the strategy of the different approaches. Methods such as SFDA-DE and CDCL are essentially based on the pseudo-labels obtained by a clustering method. Indeed, once the pseudo-labels are determined, they are pushed near the anchor (CDCL) or estimated source features (SFDA-DE). The ability of the clustering method to produce accurate pseudo-labels is essential to make these methods efficient. However, by using the \becl model on CAMELYON for the source produced at the first epoch an accuracy of 37\% on GlaS. As a result, the pseudo-labels obtained at the beginning are not reliable. Too much error is added in the early step of the process. 
This problem is also involved in SHOT but reduced by the use of an entropy loss function. This loss is a key element to consider in this context. Clustering methods can perform poorly but it does not necessarily mean that the source model is unable to produce correct output as observed in Table~\ref{tab:accuracy-bcl-source} with Deep MIL. Consequently, the error added in the adaptation process is smoothly resolved in SHOT by ensuring the same output of the method in the early stage of the training.

\section{Conclusion}
In this paper, we analyze the effectiveness of 4 representative SFDA methods for WSOL in histology images. Results indicate that SFDA can be very challenging and limited for a large dataset such as CAMELYON. Moreover, localization performance remains a challenge since SFDA methods are designed to optimize for discriminant classification. Selecting \beloc instead of \becl can lead to improvement in localization, but this incurs a decline in classification. 
Selecting \beloc for the source model doesn't imply a better localization after adaptation.

\section*{Acknowledgements}
This research was supported in part by the Canadian Institutes of Health Research, the Natural Sciences and Engineering Research Council of Canada, and the Digital Research Alliance of Canada.

\clearpage
\newpage

\appendices

\section{Hyperparameters for the source and target models}

\begin{table}[h!]
\renewcommand{\arraystretch}{1.3}
\caption{General hyper-parameters.}
\label{tab:tabx-general-hyper-params}
\centering
\resizebox{\linewidth}{!}{
\begin{tabular}{lccc}
    Hyper-parameter  &  Value  \\
    \toprule
    Fully sup. model f & U-Net\\
    \hline
    Backbones & ResNet50.\\
    \hline
    Optimizer &  SGD\\
    \hline
    Nesterov acceleration & True\\
    \hline
    \multirow{2}{*}{Momentum} & ${\in \{0.1, 0.4, 0.9\}}$ for source models\\
    &$\{ 0.4, 0.9\}$ for target models \\
    \hline
    Weight decay & $0.0001$\\
    \hline
    \multirow{2}{*}{Learning rate} & \makecell{${\in \{0.0001, 0.001, 0.002, 0.01, 0.02, 0.1\}}$} for source\\
    &$\{ 0.0001, 0.001, 0.01\}$ for target\\
    \hline
    \multirow{2}{*}{Learning rate decay} 
     & \glas: 0.1 ${\in  \{150,250,350e  \}}$ epochs for source and 250 for target\\
     & \camsixteen: 0.1 ${\in  \{2,5,8 \}}$ epochs for source and 5 for target\\
     \hline
    Mini-batch size & \makecell{$32$} \\
    \hline
    Random flip & Horizontal/vertical random flip \\
    \hline
    \multirow{2}{*}{Random color jittering} 
    & $\mathrm{Brightness}$, $\mathrm{contrast}$, \\
    & and $\mathrm{saturation}$ at $0.5$ and $\mathrm{hue}$ at $0.05$ \\
    \hline
    \multirow{2}{*}{Image size} 
    & Resize image to $225\times 225$. \\
    & Then, crop random patches of $224\times 224$\\
    \hline
    Learning epochs & \makecell{\glas: $1000$, \camsixteen: $20$} \\
    \bottomrule
\end{tabular}
}
\end{table}

\begin{table}[h!]
\renewcommand{\arraystretch}{1.3}
\caption{(WSOL) hyper-parameters.}
\label{tab:tabx-per-method-hyper-params}
\centering
\resizebox{\linewidth}{!}{
\begin{tabular}{lccc}
    Hyper-parameter  &  Value  \\
    \toprule
    Deep MIL & Mid-channels = 128. Gated attention: True/False.\\
    \bottomrule
\end{tabular}
}
\end{table}

\begin{table}[h!]
\renewcommand{\arraystretch}{1.3}
\caption{Per-method hyper-parameters for SFDA methods.}
\label{tab:tabx-per-method-hyper-params-target}
\centering
\resizebox{\linewidth}{!}{
\begin{tabular}{lccc}
    Hyper-parameter  &  Value  \\
    \toprule
    \multirow{3}{*}{SHOT} 
            & ${\lambda_{ent} \in \{0.3, 0.6, 1.0\}}$ \\
            & ${\lambda_{div} \in \{0.3, 0.6, 1.0\}}$\\
            & ${\lambda_{pseu} \in \{0.1\}}$\\
    \hline
    \multirow{2}{*}{CDCL} 
            & ${\tau \in \{0.04, 0.08\}}$ \\
            & ${\lambda_{cdcl} \in \{0.001, 0.005, 0.01, 0.05, 0.1, 1.0\}}$\\
    \hline
     \multirow{2}{*}{SFDE} 
            & ${\tau \in \{0.8, 1.0\}}$ \\
            & ${\lambda_{sfde} \in \{0.001, 0.01, 0.05, 0.1, , 0.5, 1.0\}}$\\
    \hline
    \multirow{3}{*}{AdaDSA} 
            & ${\lambda_{ent} \in \{0.03, 0.6, 1.0\}}$\\
            & ${\lambda_{ce} \in \{0.03, 0.6, 1.0\}}$\\
            & ${\lambda_{cd,smooth} \in \{0.01, 0.1\}}$\\
            & ${\alpha \in \{5,10\}}$\\
            
    \bottomrule
\end{tabular}
}
\end{table}

\section{Samples Localization Generated by our Models}
\begin{figure}[!htb]
    \centering
    \includegraphics[width=\linewidth]{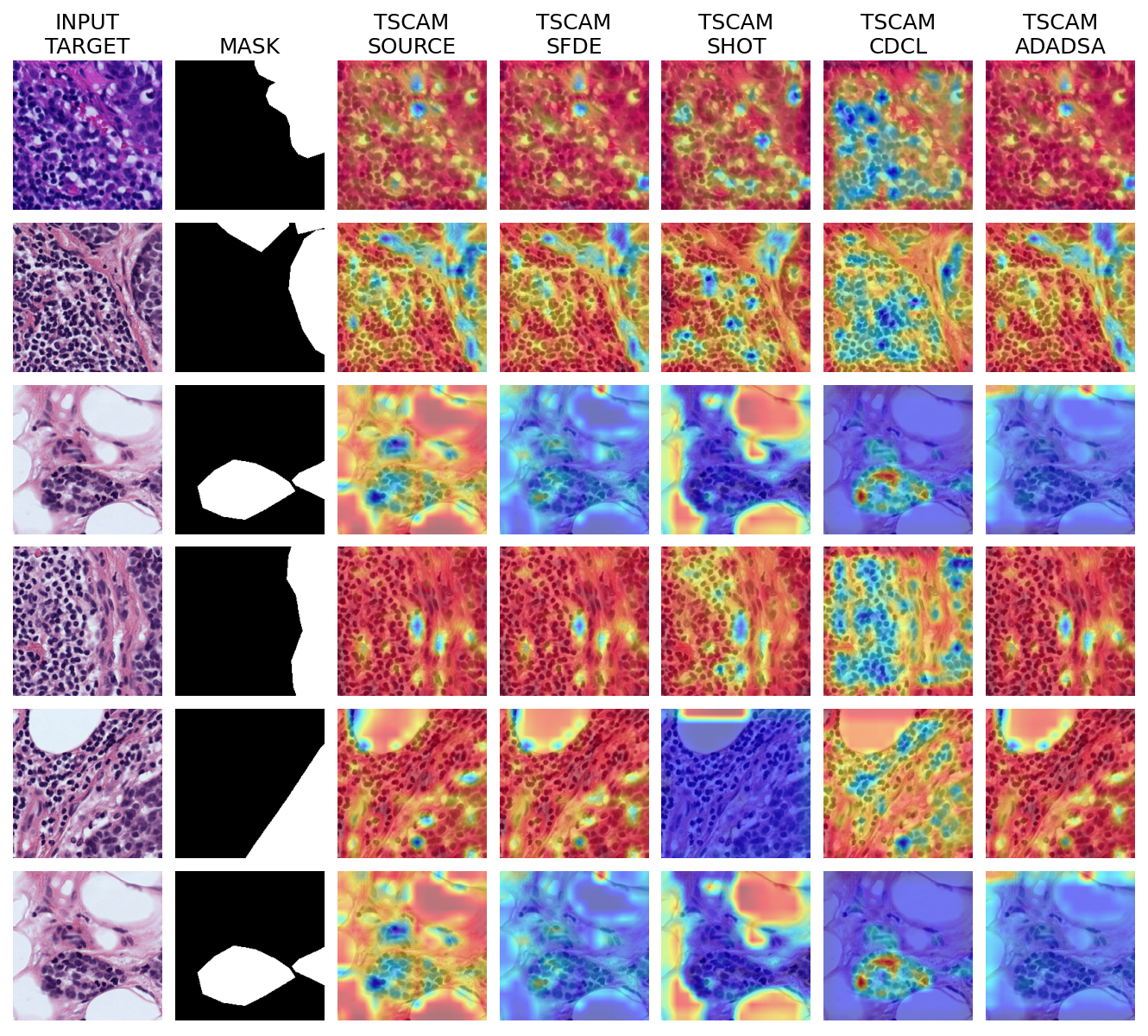}
    \caption{TSCAM best classification on CAMELYON512 without source's best classification.}
\end{figure}

\begin{figure}[!htb]
    \centering
    \includegraphics[width=\linewidth]{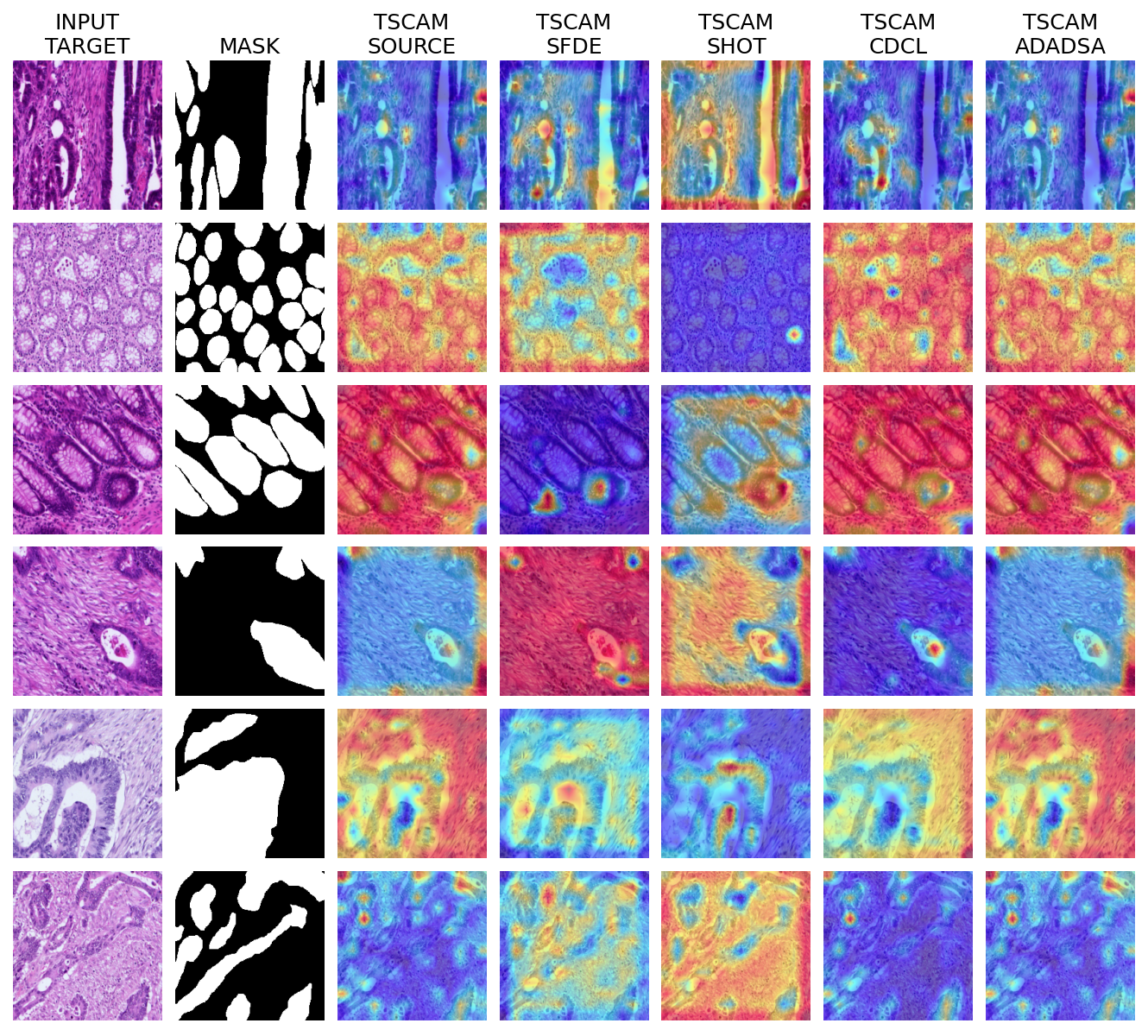}
    \caption{TSCAM best classification on GLAS with source's best classification.}
\end{figure}

\begin{figure}[!htb]
    \centering
    \includegraphics[width=\linewidth]{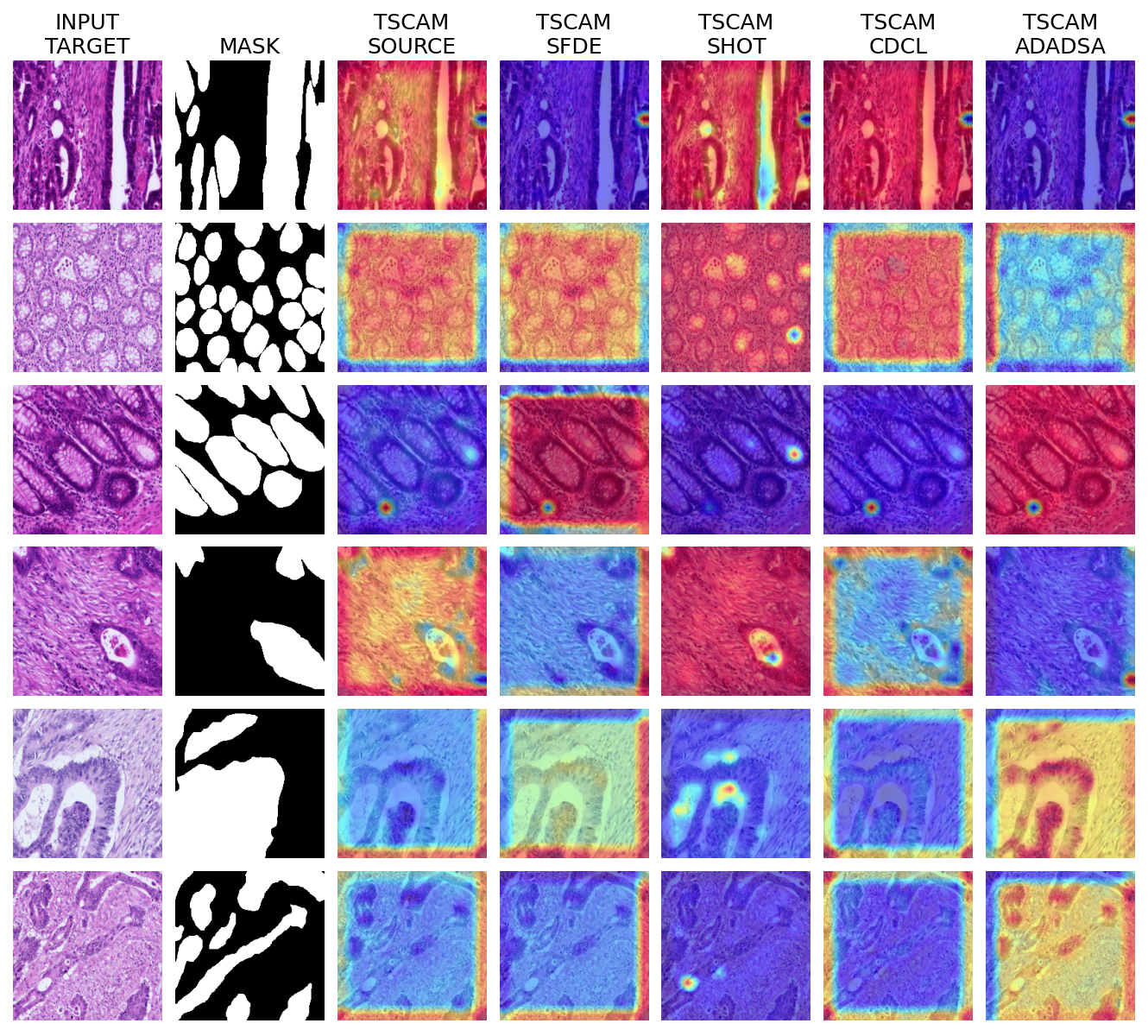}
    \caption{TSCAM best classification on GLAS without source's best classification.}
\end{figure}

\begin{figure}[!htb]
    \centering
    \includegraphics[width=\linewidth]{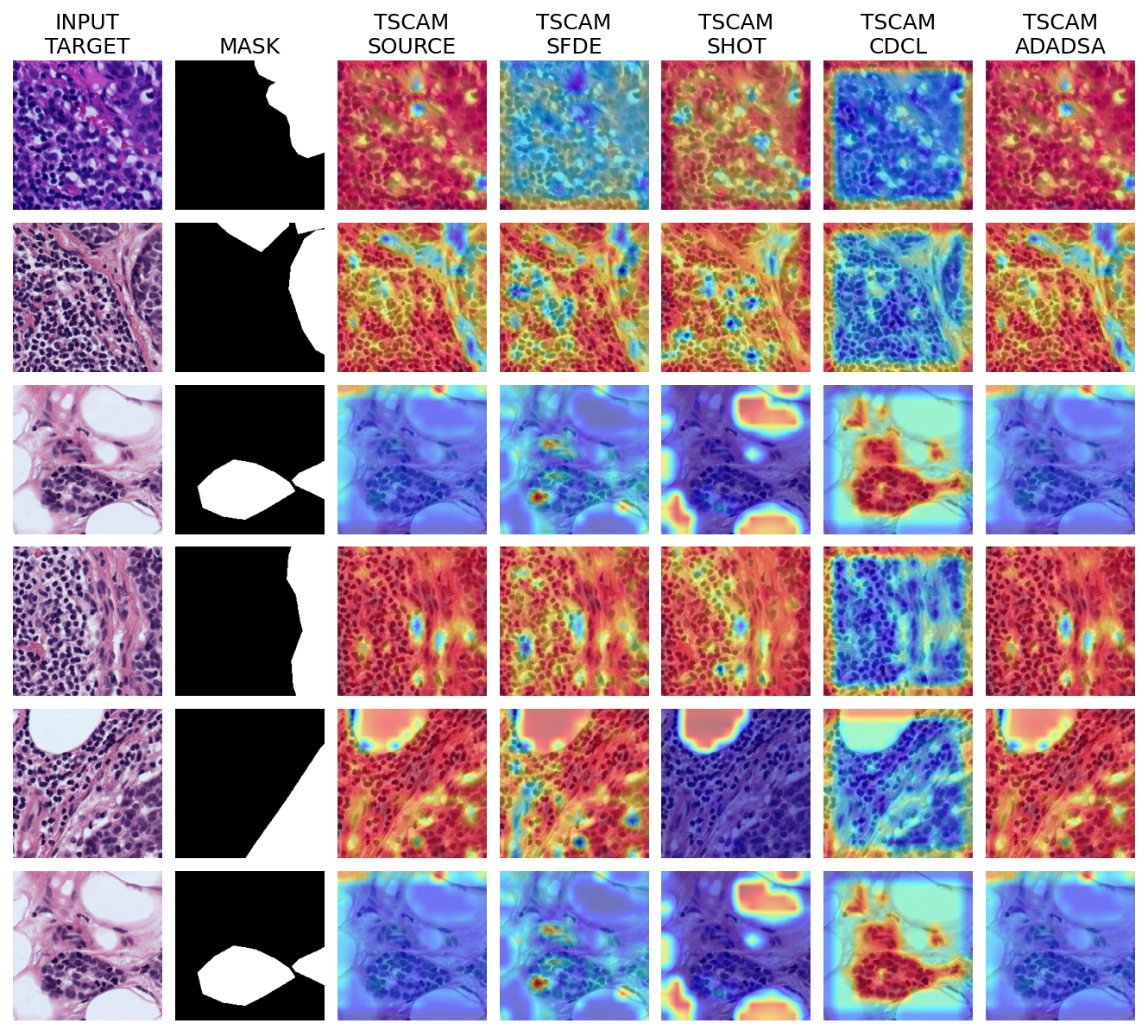}
    \caption{TSCAM best localization on CAMELYON512 without source's best classification.}
\end{figure}

\begin{figure}[!htb]
    \centering
    \includegraphics[width=\linewidth]{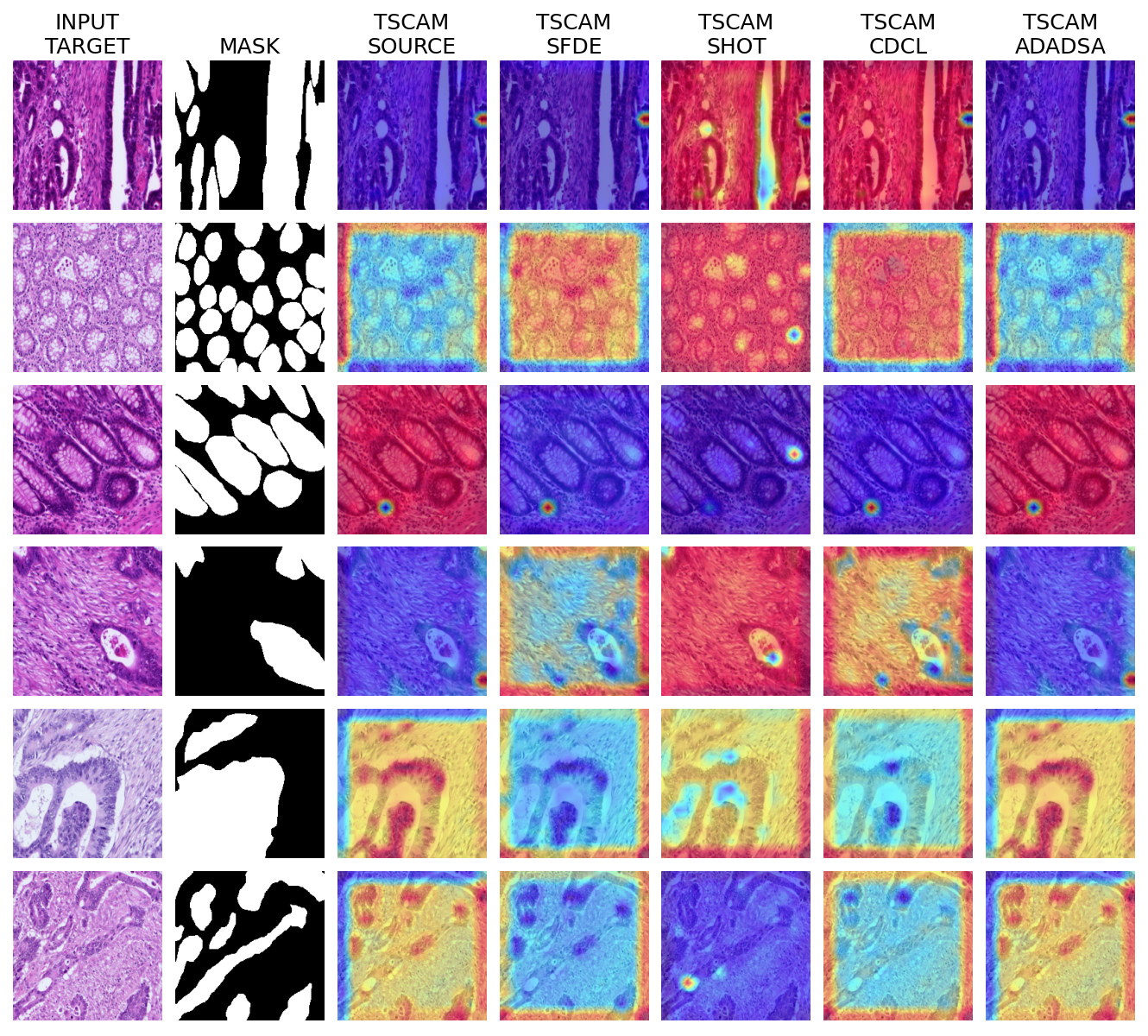}
    \caption{TSCAM best localization on GLAS without source's best classification.}
\end{figure}

\begin{figure}[!htb]
    \centering
    \includegraphics[width=\linewidth]{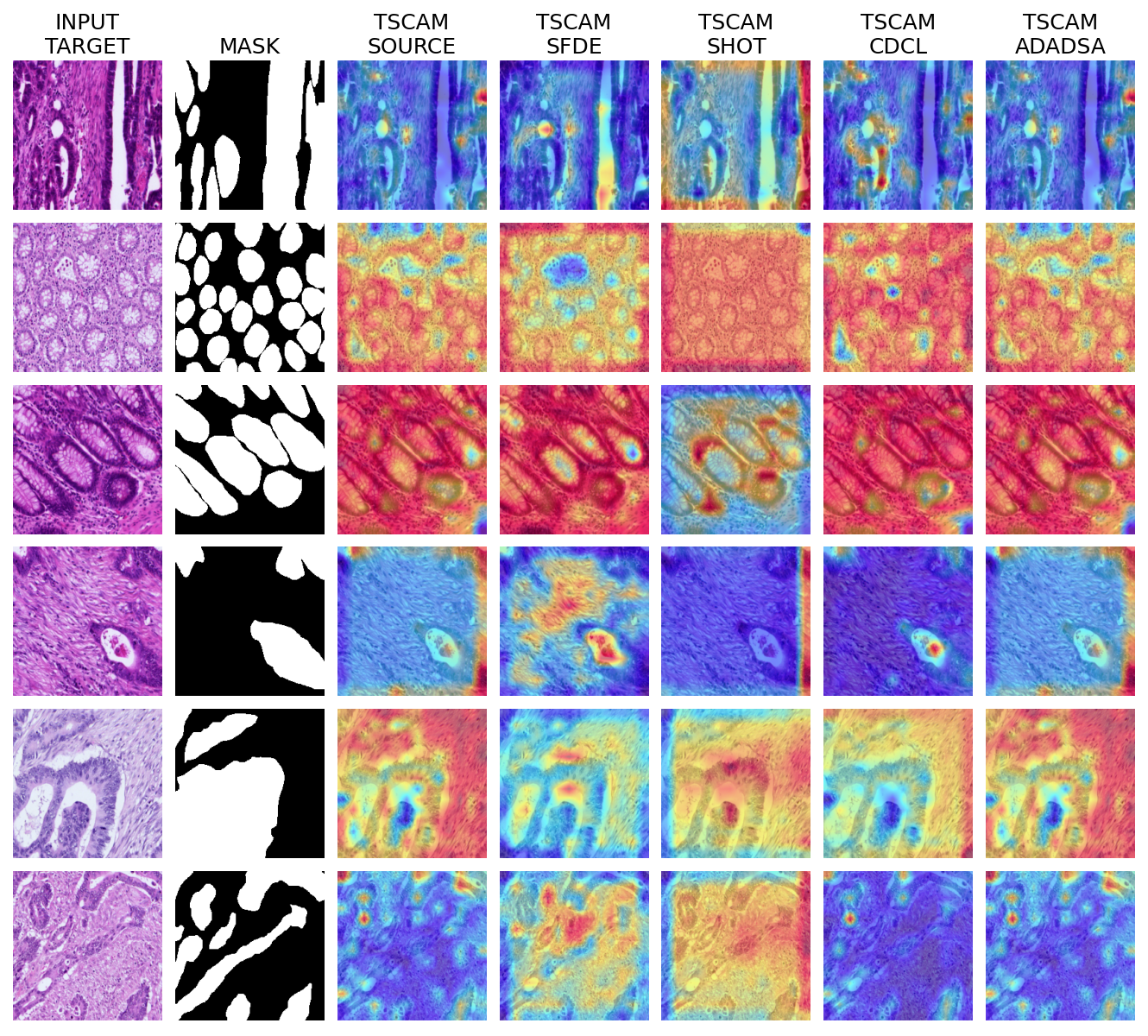}
    \caption{TSCAM best localization on GLAS with source's best classification.}
\end{figure}

\begin{figure}[!htb]
    \centering
    \includegraphics[width=\linewidth]{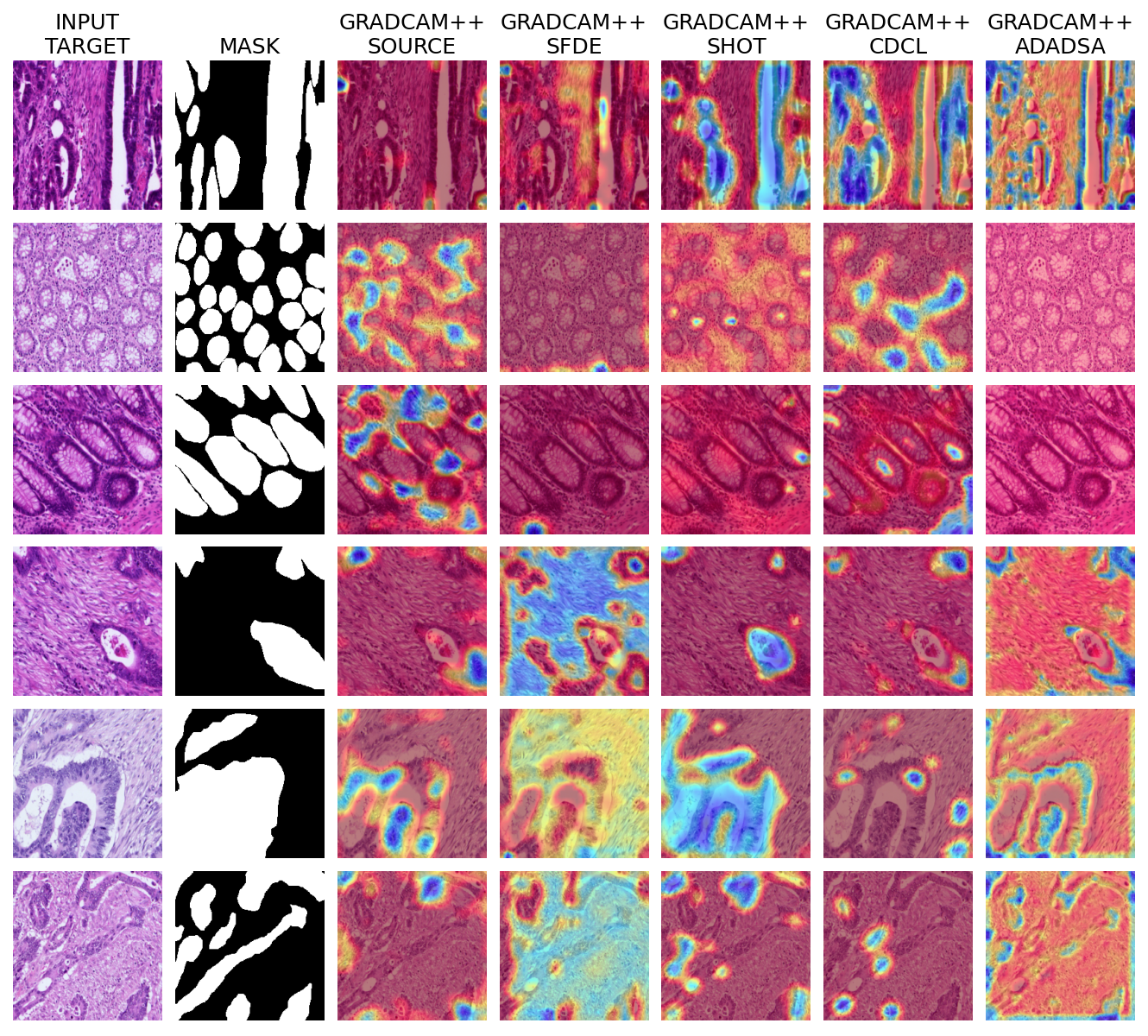}
    \caption{GradCAM++ best classification on GLAS without source's best classification.}
\end{figure}

\begin{figure}[!htb]
    \centering
    \includegraphics[width=\linewidth]{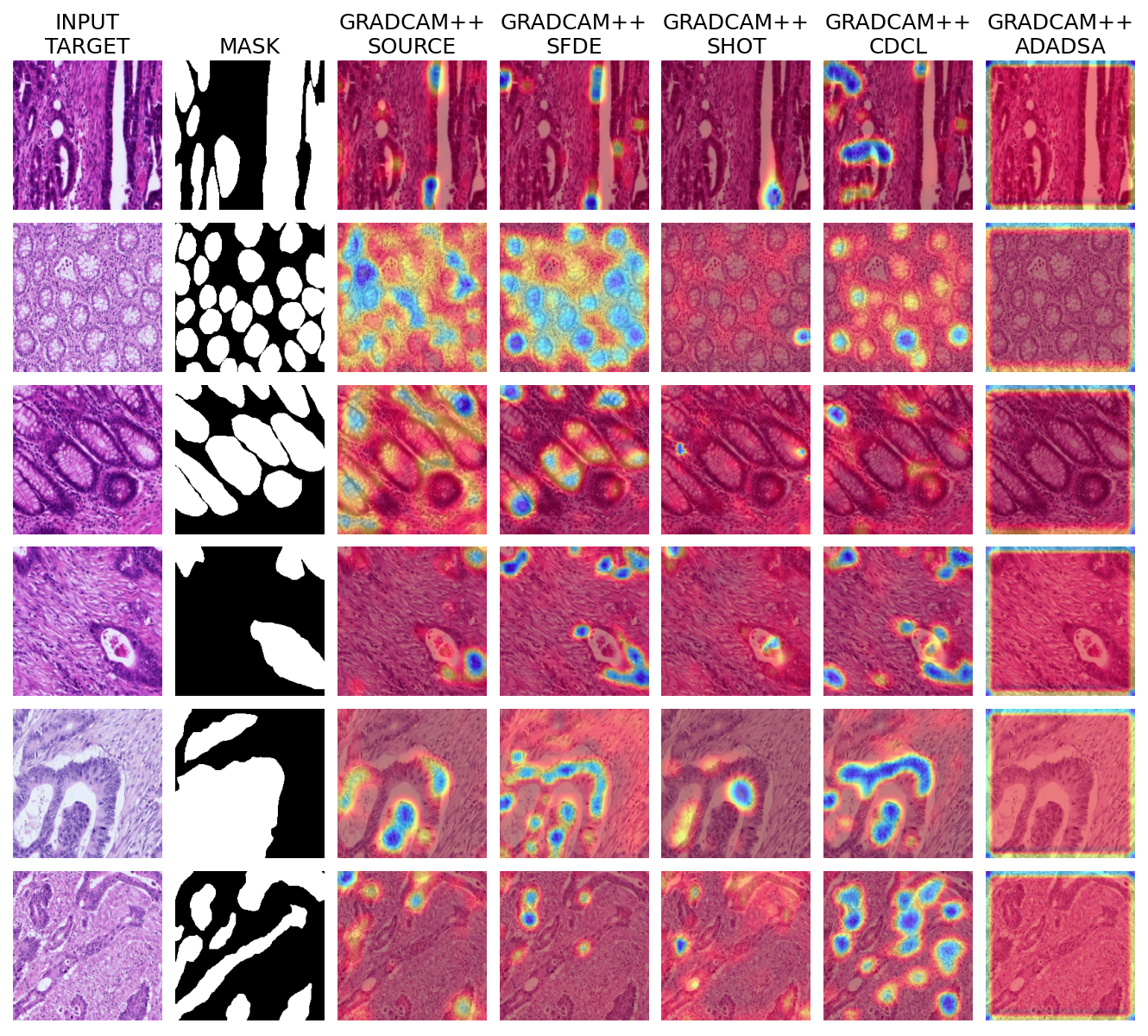}
    \caption{GradCAM++ best classification on GLAS with source's best classification.}
\end{figure}

\begin{figure}[!htb]
    \centering
    \includegraphics[width=\linewidth]{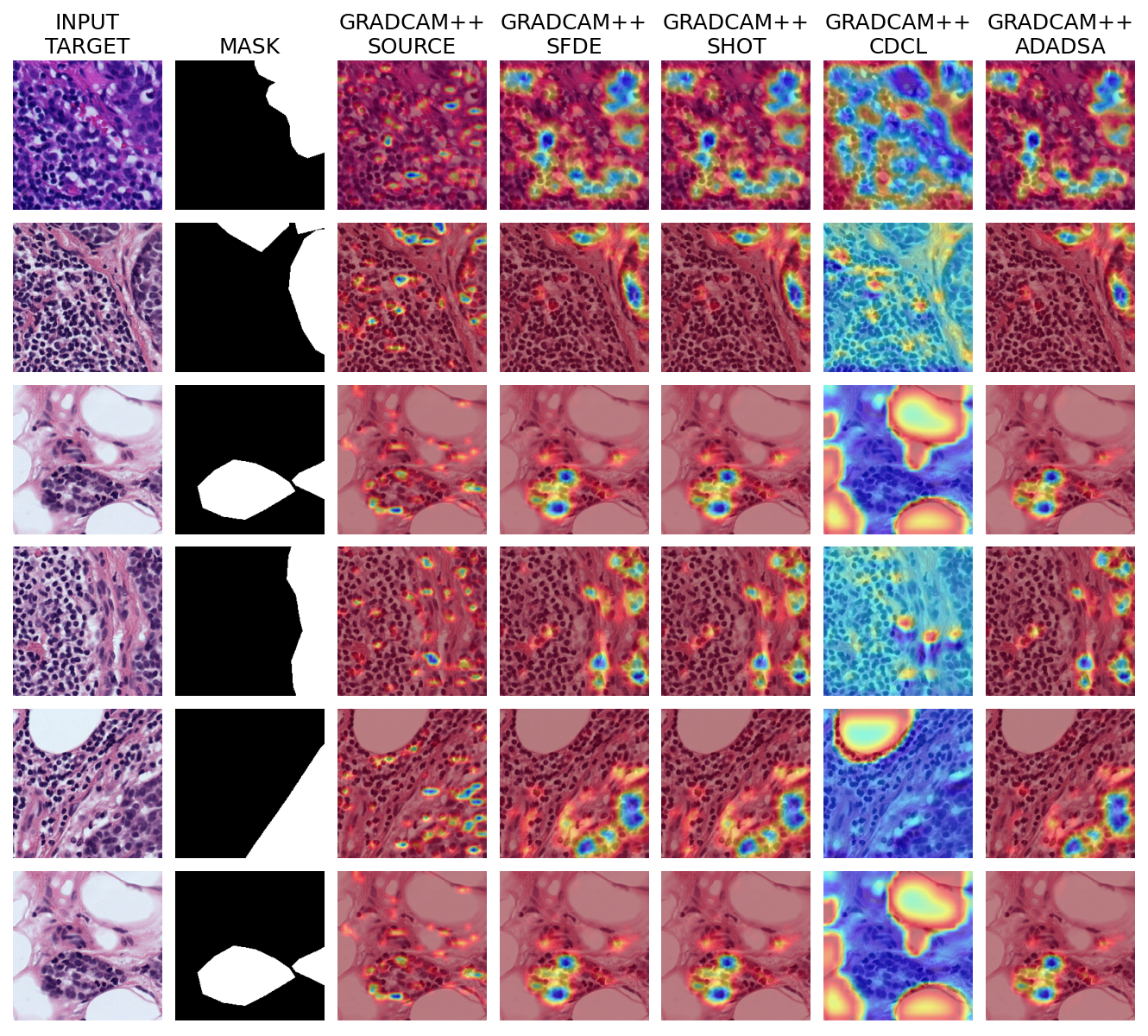}
    \caption{GradCAM++ best classification on CAMELYON512 without source's best classification.}
\end{figure}

\begin{figure}[!htb]
    \centering
    \includegraphics[width=\linewidth]{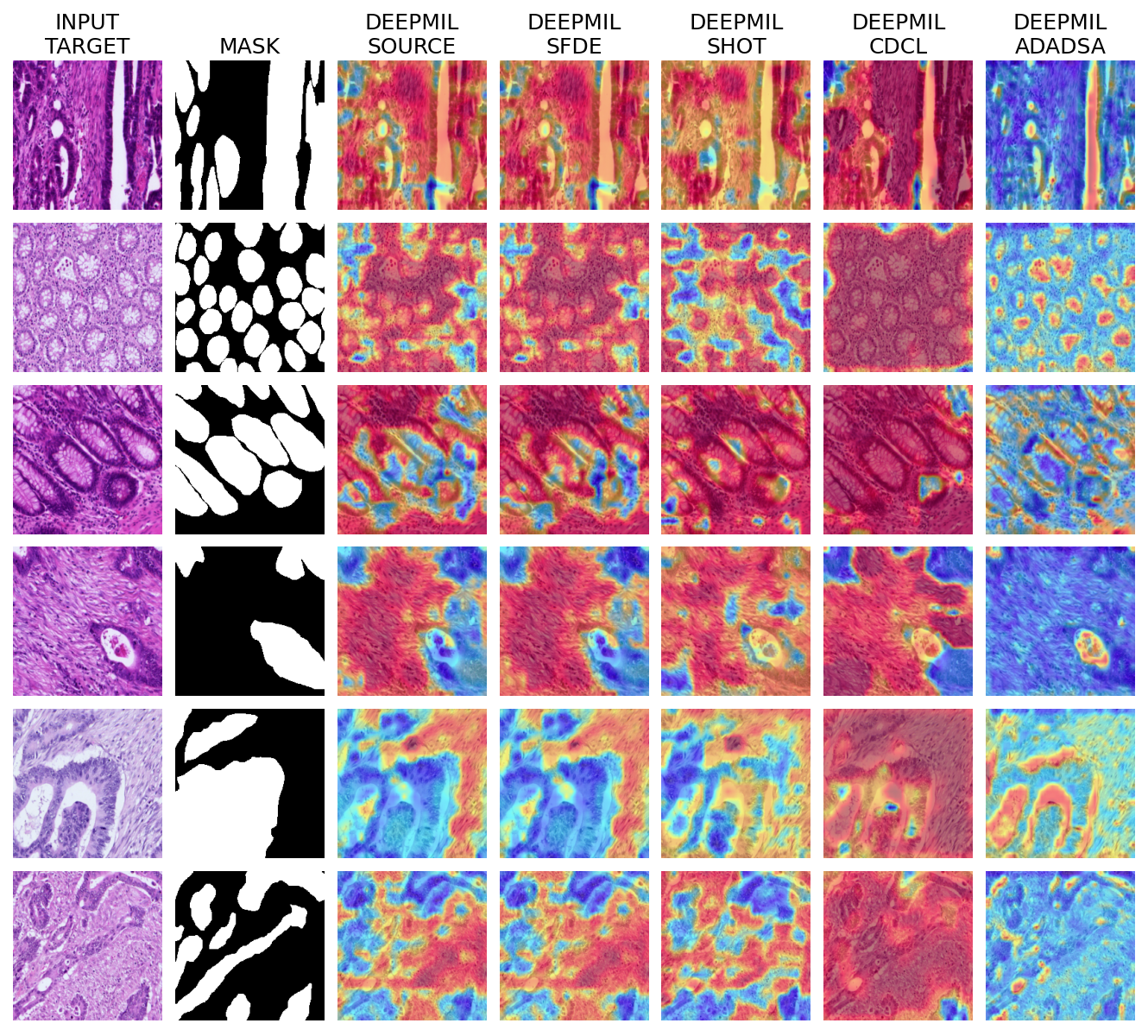}
    \caption{DEEPMIL best classification on GLAS without source's best classification.}
\end{figure}

\begin{figure}[!htb]
    \centering
    \includegraphics[width=\linewidth]{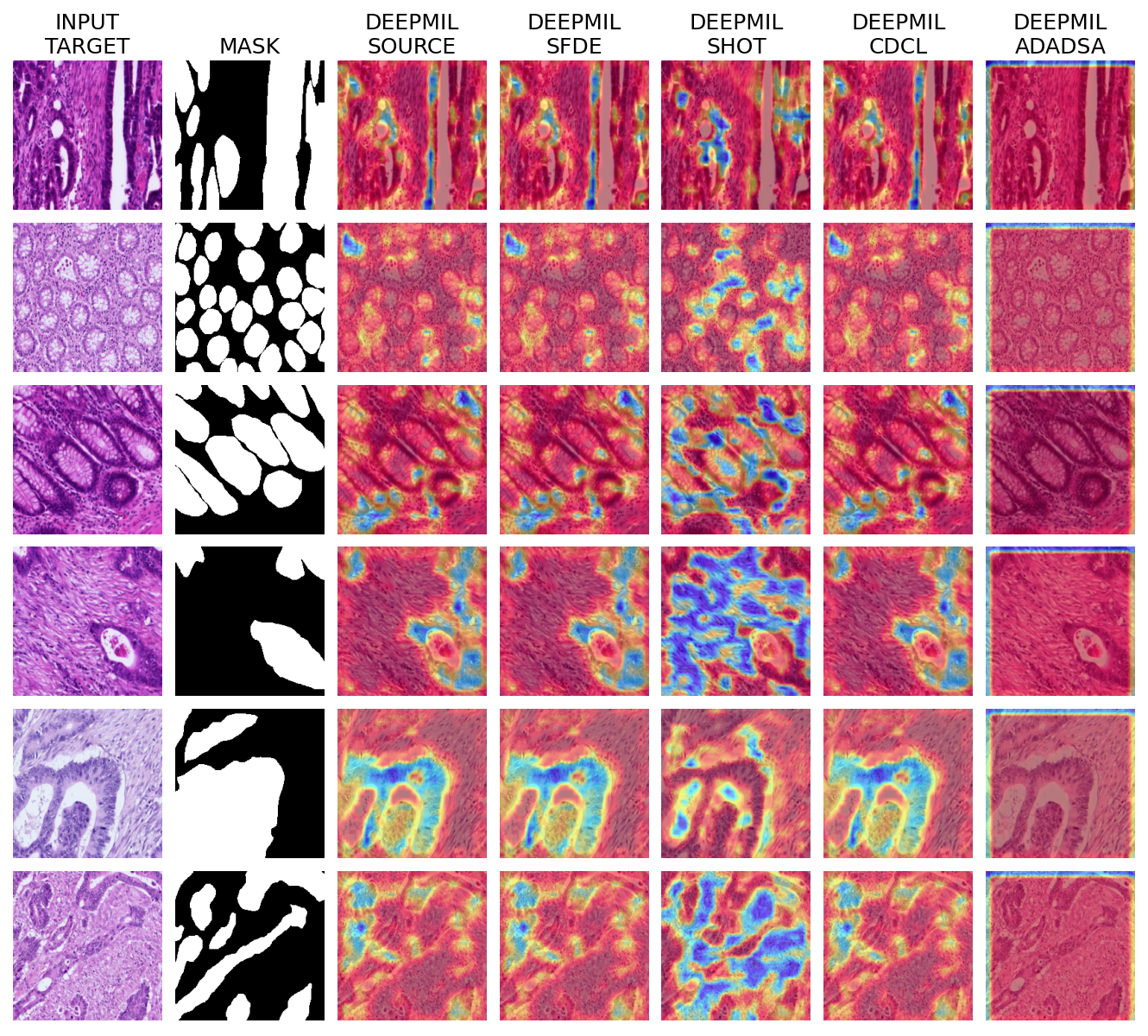}
    \caption{DEEPMIL best classification on GLAS with source's best classification.}
\end{figure}

\begin{figure}[!htb]
    \centering
    \includegraphics[width=\linewidth]{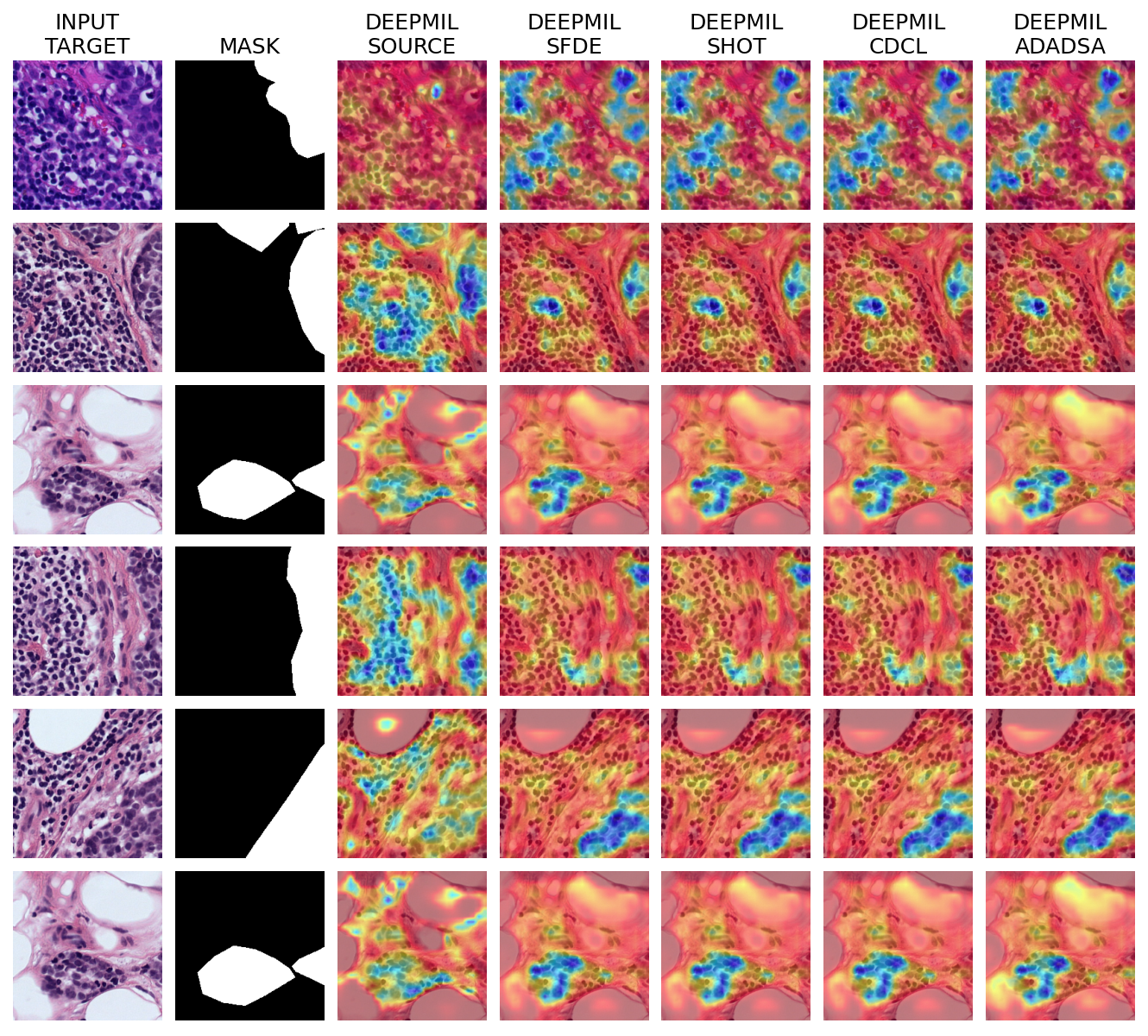}
    \caption{DEEPMIL best classification on CAMELYON512 without source's best classification.}
\end{figure}

\begin{figure}[!htb]
    \centering
    \includegraphics[width=\linewidth]{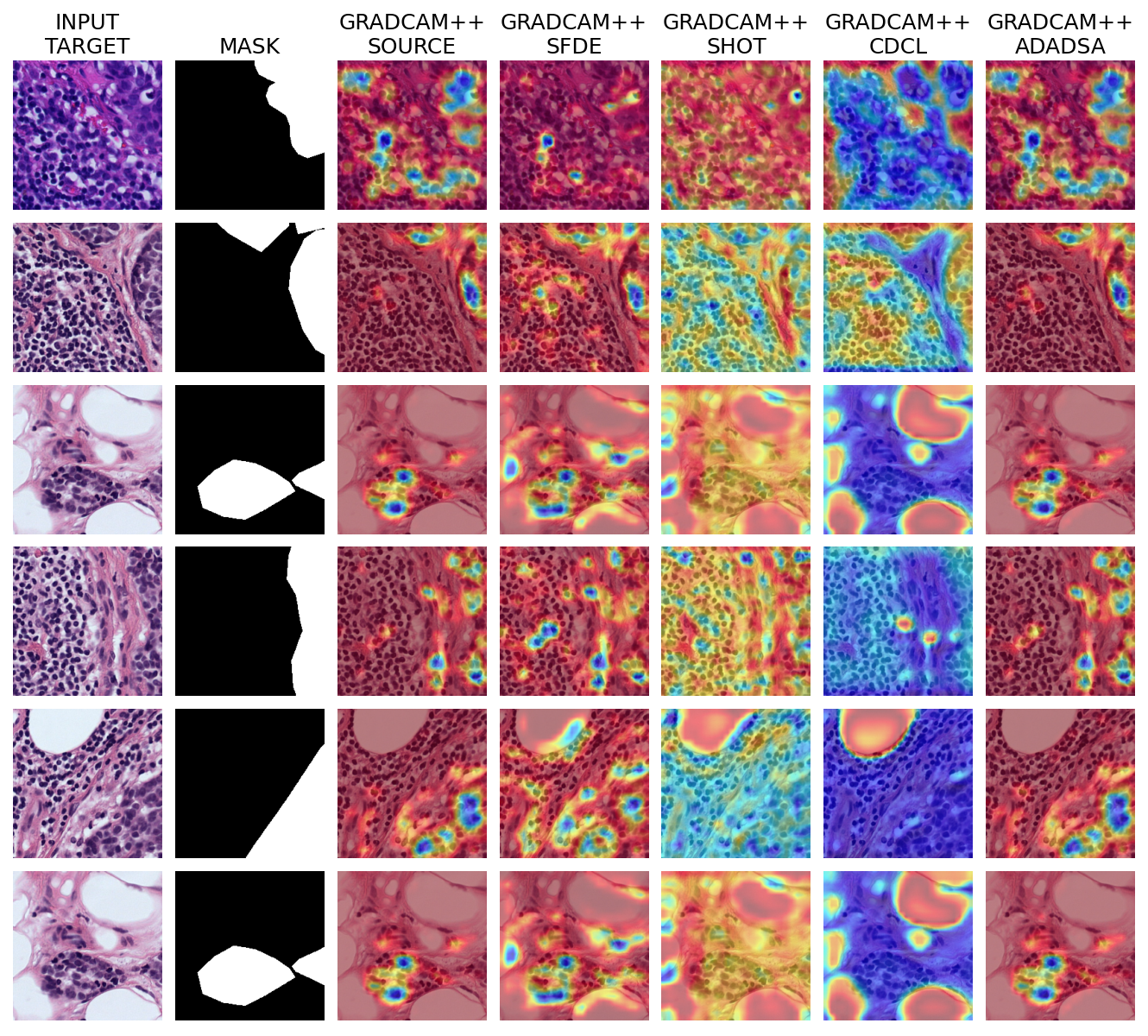}
    \caption{GradCAM++ best localization on CAMELYON512 without source's best classification.}
\end{figure}

\begin{figure}[!htb]
    \centering
    \includegraphics[width=\linewidth]{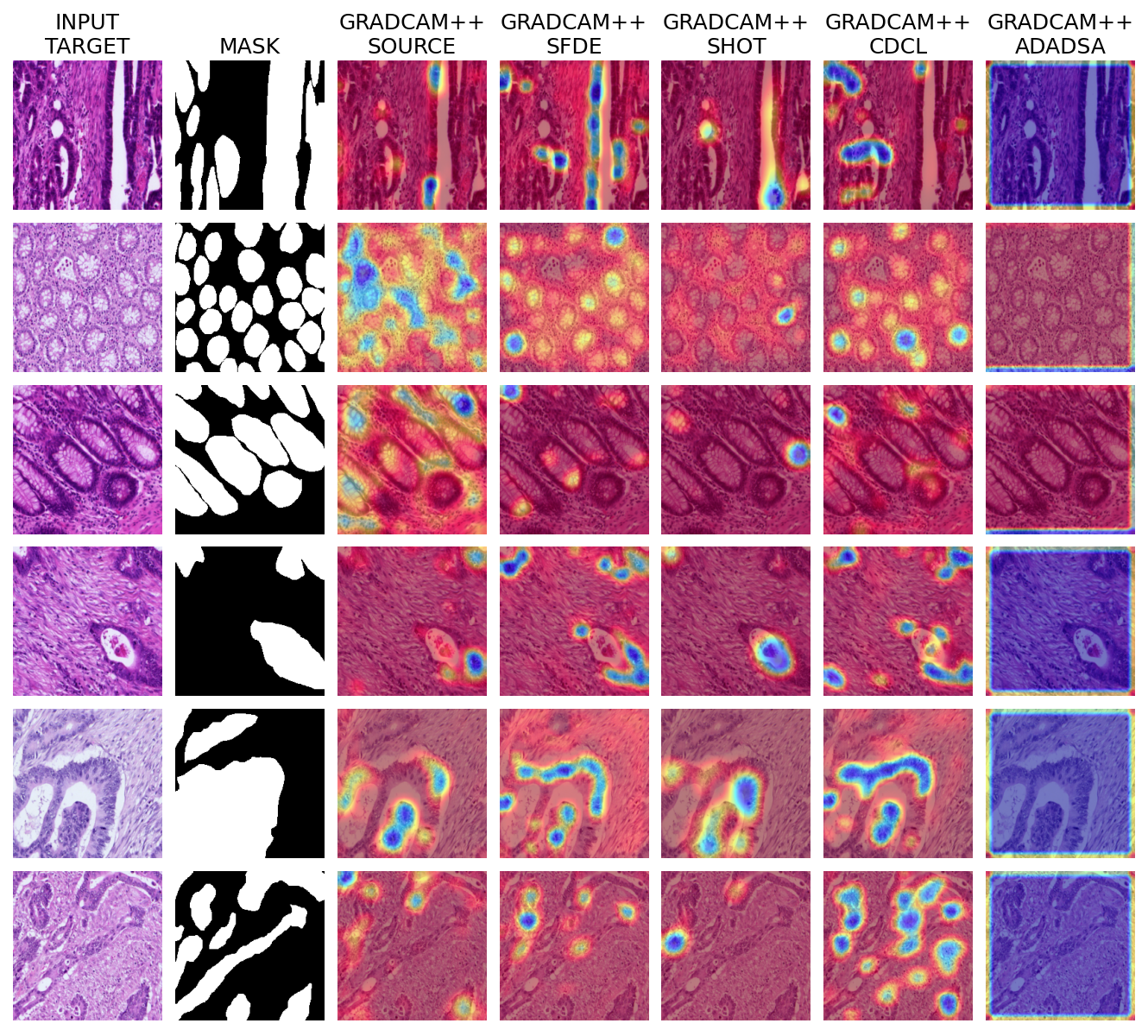}
    \caption{GradCAM++ best localization on GLAS with source's best classification.}
\end{figure}

\begin{figure}[!htb]
    \centering
    \includegraphics[width=\linewidth]{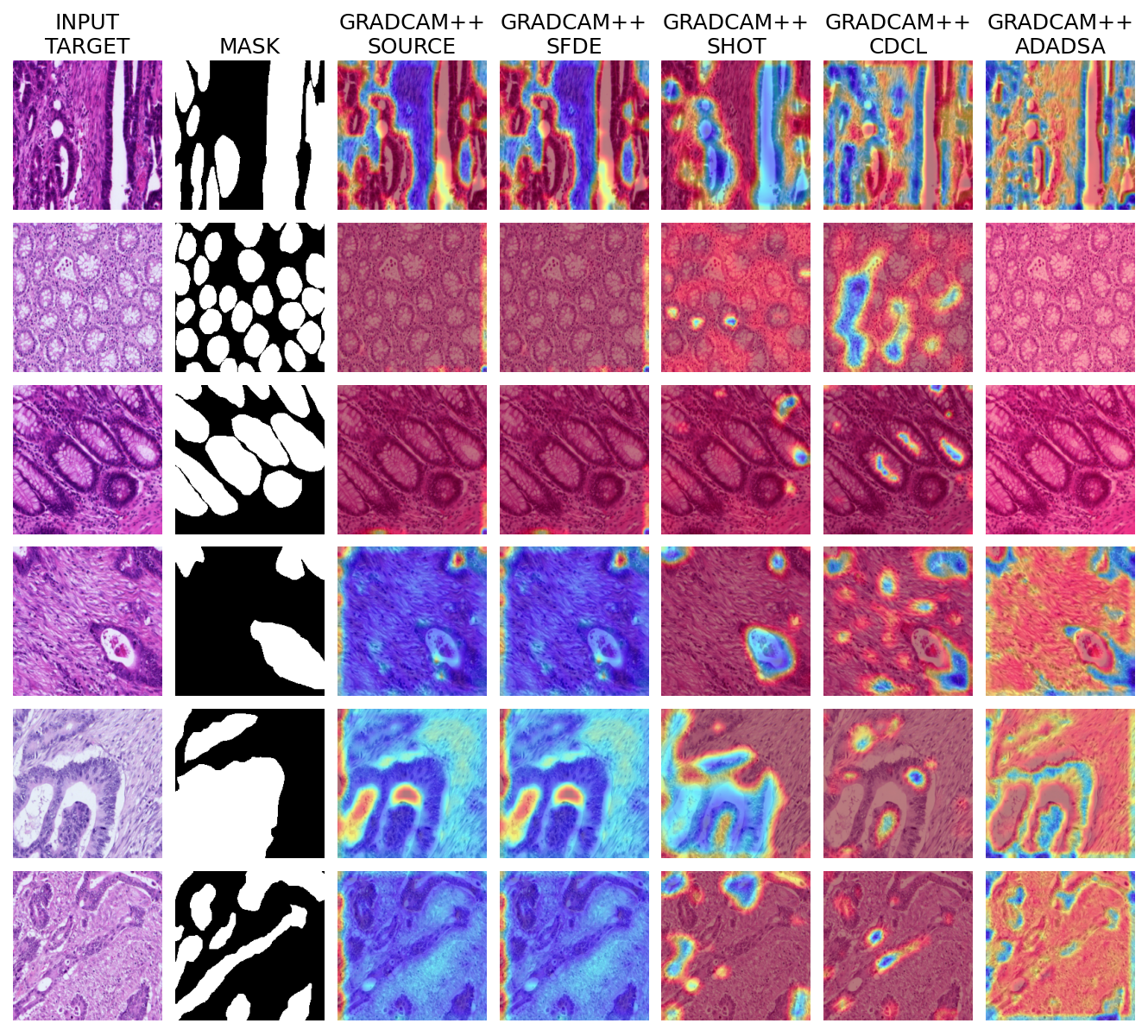}
    \caption{GradCAM++ best localization on GLAS without source's best classification.}
\end{figure}

\begin{figure}[!htb]
    \centering
    \includegraphics[width=\linewidth]{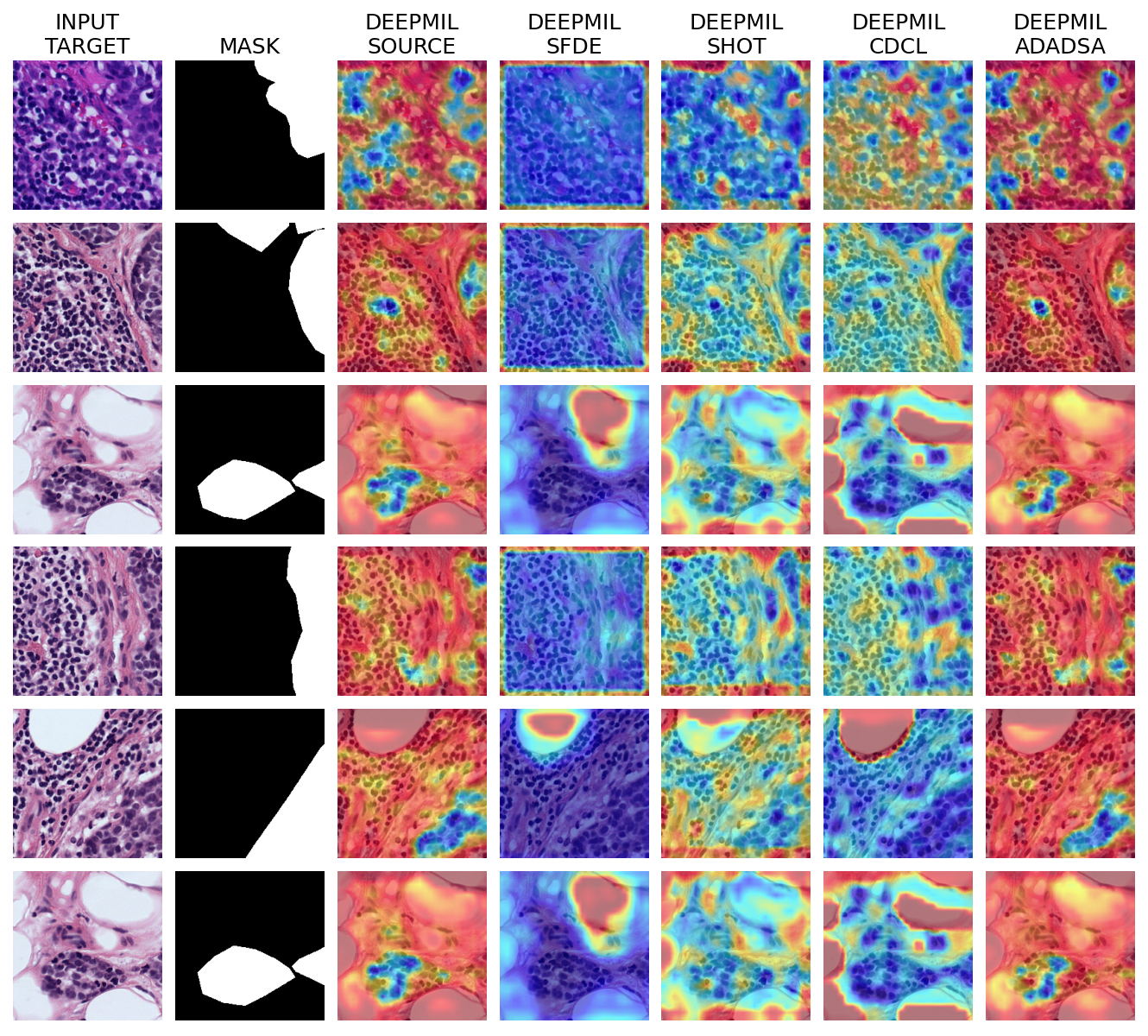}
    \caption{DEEPMIL best localization on CAMELYON512 without source's best classification.}
\end{figure}

\begin{figure}[!htb]
    \centering
    \includegraphics[width=\linewidth]{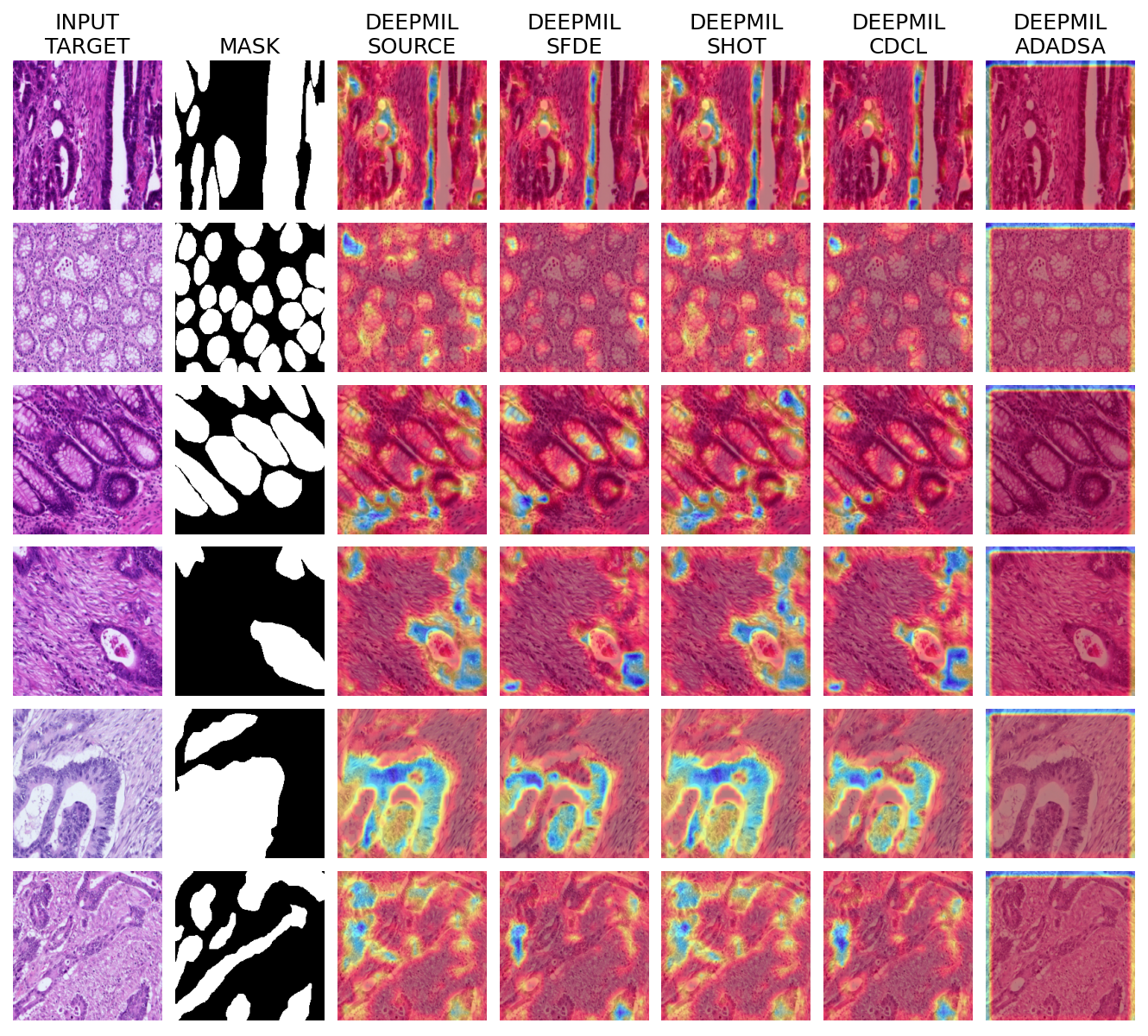}
    \caption{DEEPMIL best localization on GLAS with source's best classification.}
\end{figure}

\begin{figure}[!htb]
    \centering
    \includegraphics[width=\linewidth]{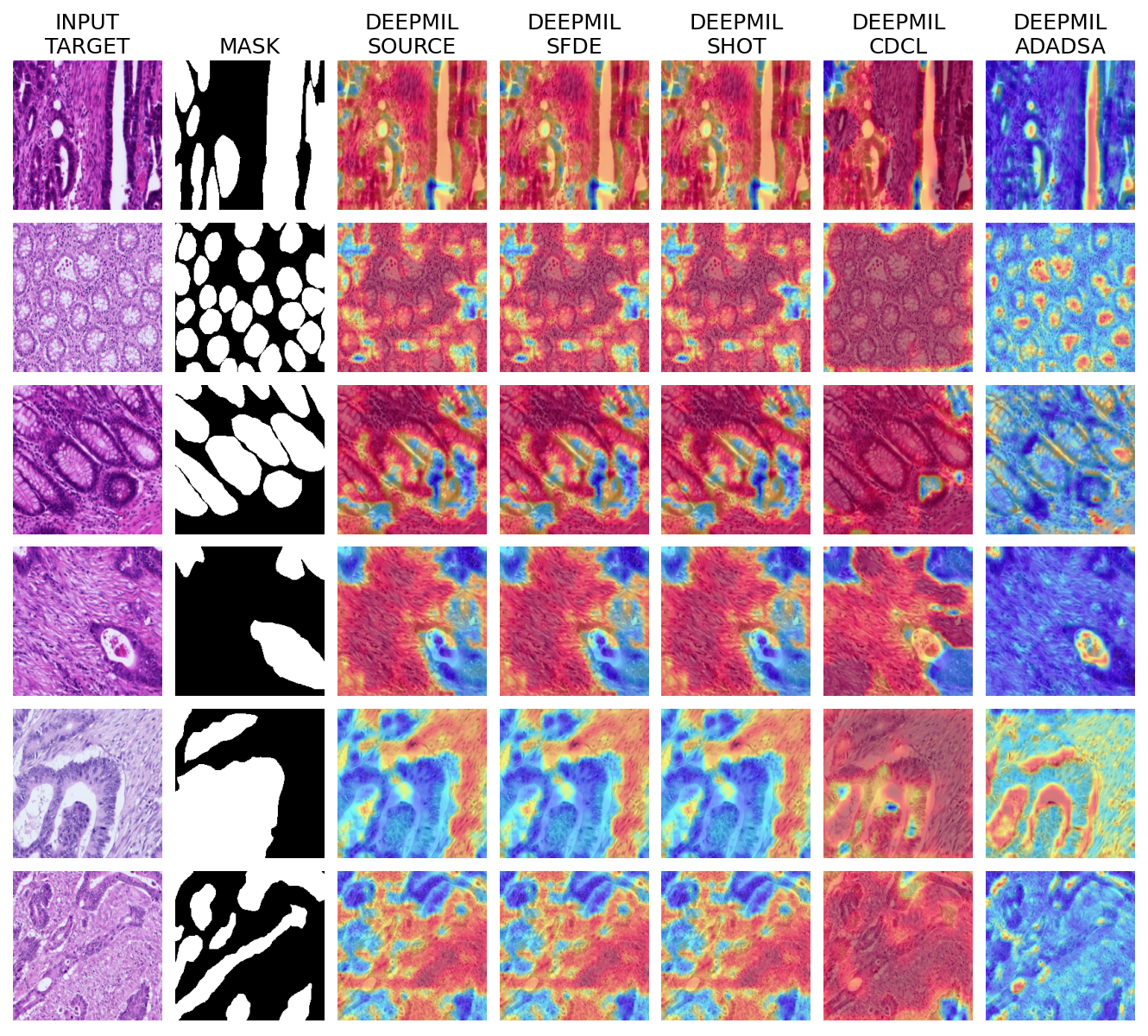}
    \caption{DEEPMIL best localization on GLAS without source's best classification.}
\end{figure}

\FloatBarrier

\bibliographystyle{apalike}
\bibliography{main}

\begin{thebibliography}{}

\bibitem[Aditya et~al., 2018]{gradcampp}
Aditya, C., Anirban, S., Prantik, H., and N, B.~V. (2018).
\newblock Grad-cam++: Generalized gradient-based visual explanations for deep convolutional networks.
\newblock In {\em WACV}.

\bibitem[Agarwal et~al., 2021]{agarwal2021unsupervised}
Agarwal, P., Paudel, D.~P., Zaech, J.-N., and Gool, L.~V. (2021).
\newblock Unsupervised robust domain adaptation without source data.
\newblock {\em CoRR}, abs/2103.14577.

\bibitem[Alex et~al., 2012]{ImageNet}
Alex, K., Ilya, S., and Hinton, G.~E. (2012).
\newblock Imagenet classification with deep convolutional neural networks.
\newblock In {\em NeurIps}.

\bibitem[Bai et~al., 2022]{bai2022weakly}
Bai, H., Zhang, R., Wang, J., and Wan, X. (2022).
\newblock Weakly supervised object localization via transformer with implicit spatial calibration.
\newblock {\em ECCV}.

\bibitem[Belharbi et~al., 2022]{belharbi2022fcam}
Belharbi, S., Sarraf, A., Pedersoli, M., Ayed, I.~B., McCaffrey, L., and Granger, E. (2022).
\newblock {F-CAM}: Full resolution class activation maps via guided parametric upscaling.
\newblock In {\em WACV}.

\bibitem[Chattopadhyay et~al., 2018]{ChattopadhyaySH18wacvgradcampp}
Chattopadhyay, A., Sarkar, A., Howlader, P., and Balasubramanian, V.~N. (2018).
\newblock Grad-cam++: Generalized gradient-based visual explanations for deep convolutional networks.
\newblock In {\em WACV}, pages 839--847.

\bibitem[Chen et~al., 2021]{chen2021selfsupervised}
Chen, W., Lin, L., Yang, S., Xie, D., Pu, S., Zhuang, Y., and Ren, W. (2021).
\newblock Self-supervised noisy label learning for source-free unsupervised domain adaptation.
\newblock {\em CoRR}, abs/2102.11614.

\bibitem[Chen et~al., 2022]{chen2022lctr}
Chen, Z., Wang, C., Wang, Y., Jiang, G., Shen, Y., Tai, Y., Wang, C., Zhang, W., and Cao, L. (2022).
\newblock {LCTR}: On awakening the local continuity of transformer for weakly supervised object localization.
\newblock In {\em AAAI}, volume~36, pages 410--418.

\bibitem[Choe and Shim, 2019]{choe2019attention}
Choe, J. and Shim, H. (2019).
\newblock Attention-based dropout layer for weakly supervised object localization.
\newblock In {\em CVPR}.

\bibitem[Ding et~al., 2022a]{ding2022sourcefree}
Ding, N., Xu, Y., Tang, Y., Xu, C., Wang, Y., and Tao, D. (2022a).
\newblock Source-free domain adaptation via distribution estimation.
\newblock {\em CVPR}, abs/2204.11257.

\bibitem[Ding et~al., 2022b]{ding2022proxymix}
Ding, Y., Sheng, L., Liang, J., Zheng, A., and He, R. (2022b).
\newblock Proxymix: Proxy-based mixup training with label refinery for source-free domain adaptation.
\newblock {\em CoRR}, abs/2205.14566.

\bibitem[Fan et~al., 2022a]{fan2022}
Fan, J., Zhu, H., Jiang, X., Meng, L., Chen, C., Fu, C., Yu, H., Dai, C., and Chen, W. (2022a).
\newblock Unsupervised domain adaptation by statistics alignment for deep sleep staging networks.
\newblock {\em IEEE TNSRE}, 30:205--216.

\bibitem[Fan et~al., 2022b]{fan2022ieee}
Fan, J., Zhu, H., Jiang, X., Meng, L., Chen, C., Fu, C., Yu, H., Dai, C., and Chen, W. (2022b).
\newblock Unsupervised domain adaptation by statistics alignment for deep sleep staging networks.
\newblock {\em IEEE TNSRE}, 30:205--216.

\bibitem[Fang et~al., 2024]{fang2024source}
Fang, Y., Yap, P.-T., Lin, W., Zhu, H., and Liu, M. (2024).
\newblock Source-free unsupervised domain adaptation: A survey.
\newblock {\em NEUNet}, page 106230.

\bibitem[Farahani et~al., 2021]{farahani2021brief}
Farahani, A., Voghoei, S., Rasheed, K., and Arabnia, H.~R. (2021).
\newblock A brief review of domain adaptation.
\newblock {\em ICDATA}, pages 877--894.

\bibitem[Fawcett, 2006]{fpr}
Fawcett, T. (2006).
\newblock An introduction to roc analysis.
\newblock {\em Pattern recognition letters}, 27(8):861--874.

\bibitem[Gao et~al., 2021]{tscam}
Gao, W., Wan, F., Pan, X., Peng, Z., Tian, Q., Han, Z., Zhou, B., and Ye, Q. (2021).
\newblock {TS-CAM:} token semantic coupled attention map for weakly supervised object localization.
\newblock In {\em ICCV}.

\bibitem[Gonz{\'{a}}lez and Woods, 2008]{woods2008}
Gonz{\'{a}}lez, R.~C. and Woods, R.~E. (2008).
\newblock {\em Digital image processing, 3rd Edition}.
\newblock Pearson Education.

\bibitem[Gupta et~al., 2022]{gupta2022vitol}
Gupta, S., Lakhotia, S., Rawat, A., and Tallamraju, R. (2022).
\newblock {ViTOL}: Vision transformer for weakly supervised object localization.
\newblock In {\em CVPRw}.

\bibitem[He et~al., 2012]{HE2012538}
He, L., Long, L.~R., Antani, S.~K., and Thoma, G.~R. (2012).
\newblock Histology image analysis for carcinoma detection and grading.
\newblock {\em CMPB}, 107(3):538--556.

\bibitem[Hegde et~al., 2021]{hegde2021uncertaintyaware}
Hegde, D., Sindagi, V., Kilic, V., Cooper, A.~B., Foster, M., and Patel, V. (2021).
\newblock Uncertainty-aware mean teacher for source-free unsupervised domain adaptive 3d object detection.
\newblock {\em CoRR}, abs/2109.14651.

\bibitem[Hou and Zheng, 2021]{hou2021source}
Hou, Y. and Zheng, L. (2021).
\newblock Source free domain adaptation with image translation.
\newblock {\em CoRR}, abs/2008.07514.

\bibitem[Ishii and Sugiyama, 2021]{ishii2021sourcefree}
Ishii, M. and Sugiyama, M. (2021).
\newblock Source-free domain adaptation via distributional alignment by matching batch normalization statistics.
\newblock {\em CoRR}, abs/2101.10842.

\bibitem[Junsuk et~al., 2020]{choe2020evaluating}
Junsuk, C., Joon, O.~S., Seungho, L., Sanghyuk, C., Zeynep, A., and Shim, H. (2020).
\newblock Evaluating weakly supervised object localization methods right.
\newblock In {\em CVPR}.

\bibitem[Kang et~al., 2019]{kang2019contrastive}
Kang, G., Jiang, L., Yang, Y., and Hauptmann, A.~G. (2019).
\newblock Contrastive adaptation network for unsupervised domain adaptation.
\newblock {\em CoRR}, abs/1901.00976.

\bibitem[Komura and Ishikawa, 2018]{komura2018machine}
Komura, D. and Ishikawa, S. (2018).
\newblock Machine learning methods for histopathological image analysis.
\newblock {\em CSBJ}, 16:34--42.

\bibitem[Kouw and Loog, 2019]{kouw2019review}
Kouw, W.~M. and Loog, M. (2019).
\newblock A review of domain adaptation without target labels.
\newblock {\em IEEE TPAMI}, 43(3):766--785.

\bibitem[Kurmi et~al., 2021]{kurmi2021domain}
Kurmi, V.~K., Subramanian, V.~K., and Namboodiri, V.~P. (2021).
\newblock Domain impression: A source data free domain adaptation method.
\newblock {\em CoRR}, abs/2102.09003.

\bibitem[Li, 2022]{li2022caft}
Li, M. (2022).
\newblock {CaFT}: Clustering and filter on tokens of transformer for weakly supervised object localization.
\newblock {\em CoRR}, abs/2201.00475.

\bibitem[Li et~al., 2020]{Li2020cvpr}
Li, R., Jiao, Q., Cao, W., Wong, H.-S., and Wu, S. (2020).
\newblock Model adaptation: Unsupervised domain adaptation without source data.
\newblock In {\em CVPR}, pages 9638--9647.

\bibitem[Liang et~al., 2020]{yang2021transformerbased}
Liang, J., Hu, D., and Feng, J. (2020).
\newblock Do we really need to access the source data? source hypothesis transfer for unsupervised domain adaptation.
\newblock {\em ICML}.

\bibitem[Liu et~al., 2021]{liu2021miccai}
Liu, X., Xing, F., Yang, C., El~Fakhri, G., and Woo, J. (2021).
\newblock Adapting off-the-shelf source segmenter for target medical image segmentation.
\newblock In {\em MICCAI 2021}, pages 549--559, Cham. Springer International Publishing.

\bibitem[Liu and Yuan, 2022]{liu2022ieeets}
Liu, X. and Yuan, Y. (2022).
\newblock A source-free domain adaptive polyp detection framework with style diversification flow.
\newblock {\em IEEE TMI}, 41(7):1897--1908.

\bibitem[Liu and Zhang, 2021]{liu2021graph}
Liu, X. and Zhang, S. (2021).
\newblock Graph consistency based mean-teaching for unsupervised domain adaptive person re-identification.
\newblock {\em CoRR}, abs/2105.04776.

\bibitem[Maximilian et~al., 2018]{deepmil}
Maximilian, I., Jakub, T., and Max, W. (2018).
\newblock Attention-based deep multiple instance learning.
\newblock In {\em ICML}.

\bibitem[Meng et~al., 2021]{meng2021foreground}
Meng, M., Zhang, T., Tian, Q., Zhang, Y., and Wu, F. (2021).
\newblock Foreground activation maps for weakly supervised object localization.
\newblock In {\em ICCV}, pages 3385--3395.

\bibitem[Meng et~al., 2022]{meng2022adversarial}
Meng, M., Zhang, T., Zhang, Z., Zhang, Y., and Wu, F. (2022).
\newblock Adversarial transformers for weakly supervised object localization.
\newblock {\em IEEE TIP}, 31:7130--7143.

\bibitem[Murtaza et~al., 2022a]{murtaza2022dipssypo}
Murtaza, S., Belharbi, S., Pedersoli, M., Sarraf, A., and Granger, E. (2022a).
\newblock Constrained sampling for class-agnostic weakly supervised object localization.
\newblock In {\em Montreal AI symposium}.

\bibitem[Murtaza et~al., 2022b]{murtaza2022dips}
Murtaza, S., Belharbi, S., Pedersoli, M., Sarraf, A., and Granger, E. (2022b).
\newblock Discriminative sampling of proposals in self-supervised transformers for weakly supervised object localization.
\newblock {\em CoRR}, abs/2209.09209.

\bibitem[Murtaza et~al., 2023]{Murtaza2023dips}
Murtaza, S., Belharbi, S., Pedersoli, M., Sarraf, A., and Granger, E. (2023).
\newblock Dips: Discriminative pseudo-label sampling with self-supervised transformers for weakly supervised object localization.
\newblock {\em IMAVIS}, page 104838.

\bibitem[Pan et~al., 2021]{pan2021unveiling}
Pan, X., Gao, Y., Lin, Z., Tang, F., Dong, W., Yuan, H., Huang, F., and Xu, C. (2021).
\newblock Unveiling the potential of structure preserving for weakly supervised object localization.
\newblock In {\em CVPR}, pages 11642--11651.

\bibitem[Qiu et~al., 2021]{qiu2021sourcefree}
Qiu, Z., Zhang, Y., Lin, H., Niu, S., Liu, Y., Du, Q., and Tan, M. (2021).
\newblock Source-free domain adaptation via avatar prototype generation and adaptation.
\newblock {\em CoRR}, abs/2106.15326.

\bibitem[Rahimi et~al., 2020]{rahimi2020pairwise}
Rahimi, A., Shaban, A., Ajanthan, T., Hartley, R., and Boots, B. (2020).
\newblock Pairwise similarity knowledge transfer for weakly supervised object localization.
\newblock In {\em ECCV}.

\bibitem[Ronneberger et~al., 2015]{unet}
Ronneberger, O., Fischer, P., and Brox, T. (2015).
\newblock U-net: Convolutional networks for biomedical image segmentation.
\newblock In {\em MICCAI}.

\bibitem[Rony et~al., 2023]{rony23}
Rony, J., Belharbi, S., Dolz, J., Ben~Ayed, I., McCaffrey, L., and Granger, E. (2023).
\newblock Deep weakly-supervised learning methods for classification and localization in histology images: A survey.
\newblock {\em MLBI}, 2:96--150.

\bibitem[Sahoo et~al., 1988]{Thresholding_techniques}
Sahoo, P., Soltani, S., and Wong, A. (1988).
\newblock A survey of thresholding techniques.
\newblock {\em CVGIP}, 41(2):233--260.

\bibitem[Selvaraju et~al., 2017]{SelvarajuCDVPB17iccvgradcam}
Selvaraju, R.~R., Cogswell, M., Das, A., Vedantam, R., Parikh, D., and Batra, D. (2017).
\newblock Grad-cam: Visual explanations from deep networks via gradient-based localization.
\newblock In {\em ICCV}, pages 618--626.

\bibitem[Singh and Lee, 2017]{singh2017hide}
Singh, K.~K. and Lee, Y.~J. (2017).
\newblock Hide-and-seek: Forcing a network to be meticulous for weakly-supervised object and action localization.
\newblock In {\em ICCV}.

\bibitem[Su et~al., 2022]{su2022re}
Su, H., Ye, Y., Chen, Z., Song, M., and Cheng, L. (2022).
\newblock Re-attention transformer for weakly supervised object localization.
\newblock In {\em BMVC}.

\bibitem[Subtypes, 2022]{subtypes2022pathology}
Subtypes, R. (2022).
\newblock Pathology of breast cancer.
\newblock {\em Updates in the Management of Breast Cancer, An Issue of Surgical Clinics}, 103(1):1.

\bibitem[Tian et~al., 2021]{tian2021vdmda}
Tian, J., Zhang, J., Li, W., and Xu, D. (2021).
\newblock Vdm-da: Virtual domain modeling for source data-free domain adaptation.
\newblock {\em CoRR}, abs/2103.14357.

\bibitem[Wang and Deng, 2018]{wang2018deep}
Wang, M. and Deng, W. (2018).
\newblock Deep visual domain adaptation: A survey.
\newblock {\em NEUCOMPU}, 312:135--153.

\bibitem[Wang et~al., 2022a]{wang2022cross}
Wang, R., Wu, Z., Weng, Z., Chen, J., Qi, G.-J., and Jiang, Y.-G. (2022a).
\newblock Cross-domain contrastive learning for unsupervised domain adaptation.
\newblock {\em IEEE TMM}.

\bibitem[Wang et~al., 2022b]{wang2022crossdomain}
Wang, R., Wu, Z., Weng, Z., Chen, J., Qi, G.-J., and Jiang, Y.-G. (2022b).
\newblock Cross-domain contrastive learning for unsupervised domain adaptation.
\newblock {\em CoRR}, abs/2106.05528.

\bibitem[Wei et~al., 2017]{wei2017object}
Wei, Y., Feng, J., Liang, X., Cheng, M., Zhao, Y., and Yan, S. (2017).
\newblock Object region mining with adversarial erasing: A simple classification to semantic segmentation approach.
\newblock In {\em CVPR}.

\bibitem[Wu et~al., 2022]{wu2021background}
Wu, P., Zhai, W., and Cao, Y. (2022).
\newblock Background activation suppression for weakly supervised object localization.
\newblock In {\em CVPR}.

\bibitem[Xia et~al., 2021]{Adversarial_contrastive}
Xia, H., Zhao, H., and Ding, Z. (2021).
\newblock Adaptive adversarial network for source-free domain adaptation.
\newblock In {\em ICCV}, pages 8990--8999.

\bibitem[Xie et~al., 2021]{xie2021online}
Xie, J., Luo, C., Zhu, X., Jin, Z., Lu, W., and Shen, L. (2021).
\newblock Online refinement of low-level feature based activation map for weakly supervised object localization.
\newblock In {\em ICCV}, pages 132--141.

\bibitem[Xue et~al., 2019]{xue2019danet}
Xue, H., Liu, C., Wan, F., Jiao, J., Ji, X., and Ye, Q. (2019).
\newblock Danet: Divergent activation for weakly supervised object localization.
\newblock In {\em CVPR}.

\bibitem[Yang et~al., 2022]{yang2022source}
Yang, C., Guo, X., Chen, Z., and Yuan, Y. (2022).
\newblock Source free domain adaptation for medical image segmentation with fourier style mining.
\newblock {\em Medical Image Analysis}, 79:102457.

\bibitem[Yang et~al., 2021a]{yang2021exploiting}
Yang, S., van~de Weijer, J., Herranz, L., Jui, S., et~al. (2021a).
\newblock Exploiting the intrinsic neighborhood structure for source-free domain adaptation.
\newblock {\em NeurIPS}, 34:29393--29405.

\bibitem[Yang et~al., 2021b]{yang2021generalized}
Yang, S., Wang, Y., Van De~Weijer, J., Herranz, L., and Jui, S. (2021b).
\newblock Generalized source-free domain adaptation.
\newblock In {\em ICCV}, pages 8978--8987.

\bibitem[Yu et~al., 2022]{yu2022sourcefree}
Yu, H., Huang, J., Liu, Y., Zhu, Q., Zhou, M., and Zhao, F. (2022).
\newblock Source-free domain adaptation for real-world image dehazing.
\newblock {\em CoRR}, abs/2207.06644.

\bibitem[Yu et~al., 2023]{yu2302comprehensive}
Yu, Z., Li, J., Du, Z., Zhu, L., and Shen, H. (2023).
\newblock A comprehensive survey on source-free domain adaptation.
\newblock {\em CoRR}, abs/2302.11803.

\bibitem[Yun et~al., 2019]{yun2019cutmix}
Yun, S., Han, D., Oh, S.~J., Chun, S., Choe, J., and Yoo, Y. (2019).
\newblock Cutmix: Regularization strategy to train strong classifiers with localizable features.
\newblock In {\em ICCV}, pages 6023--6032.

\bibitem[Zhang et~al., 2020a]{zhang2020rethinking}
Zhang, C.-L., Cao, Y.-H., and Wu, J. (2020a).
\newblock Rethinking the route towards weakly supervised object localization.
\newblock In {\em CVPR}.

\bibitem[Zhang et~al., 2023]{zhang2023source}
Zhang, N., Lu, J., Li, K., Fang, Z., and Zhang, G. (2023).
\newblock Source-free unsupervised domain adaptation: Current research and future directions.
\newblock {\em NERUCOMPU}, page 126921.

\bibitem[Zhang et~al., 2018]{zhang2018adversarial}
Zhang, X., Wei, Y., Feng, J., Yang, Y., and Huang, T. (2018).
\newblock Adversarial complementary learning for weakly supervised object localization.
\newblock In {\em CVPR}.

\bibitem[Zhang et~al., 2020b]{zhang2020inter}
Zhang, X., Wei, Y., and Yang, Y. (2020b).
\newblock Inter-image communication for weakly supervised localization.
\newblock In {\em ECCV}, pages 271--287.

\bibitem[Zhou et~al., 2016]{zhou2016learning}
Zhou, B., Khosla, A., Lapedriza, A., Oliva, A., and Torralba, A. (2016).
\newblock Learning deep features for discriminative localization.
\newblock In {\em CVPR}.

\bibitem[Zhu et~al., 2023]{zhu2021background}
Zhu, L., She, Q., Chen, Q., Meng, X., Geng, M., Jin, L., Jiang, Z., Qiu, B., You, Y., and Zhang, Y. (2023).
\newblock Background-aware classification activation map for weakly supervised object localization.
\newblock {\em IEEE TPAMI}.

\bibitem[Zhu et~al., 2022]{zhu2022weakly}
Zhu, L., She, Q., Chen, Q., You, Y., Wang, B., and Lu, Y. (2022).
\newblock Weakly supervised object localization as domain adaption.
\newblock In {\em CVPR}.

\end{thebibliography}

\end{document}